\documentclass{article}

\usepackage{microtype}
\usepackage{graphicx}
\usepackage{subcaption}
\usepackage{booktabs} 
\usepackage{arydshln}
\usepackage{multirow}
\usepackage{xcolor}
\usepackage{ulem}
\usepackage[table]{xcolor}
\usepackage{pifont}
\usepackage{makecell}

\definecolor{bestgreen}{HTML}{7DCEA0}   
\definecolor{secondgreen}{HTML}{D5F5E3} 
\definecolor{upcolor}{RGB}{0,128,128}   
\definecolor{downcolor}{RGB}{178,34,34} 
\newcommand{\best}[1]{\cellcolor{bestgreen}\textbf{#1}}
\newcommand{\secondbest}[1]{\cellcolor{secondgreen}\uline{#1}}
\newcommand{\besttext}[1]{\colorbox{bestgreen}{\textbf{#1}}}
\newcommand{\secondbesttext}[1]{\colorbox{secondgreen}{\uline{#1}}}
\usepackage{hyperref}
\hypersetup{
    colorlinks=false,
    linkbordercolor={red},
    citebordercolor={green}
}

\usepackage[preprint]{icml2026}

\usepackage{amsmath}
\usepackage{amssymb}
\usepackage{mathtools}
\usepackage{amsthm}

\usepackage[capitalize,noabbrev]{cleveref}

\theoremstyle{plain}
\newtheorem{theorem}{Theorem}[section]

\theoremstyle{definition}
\newtheorem{definition}[theorem]{Definition}

\theoremstyle{remark}

\usepackage[disable,textsize=tiny]{todonotes}

\icmltitlerunning{Toward Effective Multimodal Graph Foundation Model: A Divide-and-Conquer Based Approach}

\begin{document}

\twocolumn[
  \icmltitle{Toward Effective Multimodal Graph Foundation Model: \\A Divide-and-Conquer Based Approach}



  \icmlsetsymbol{equal}{*}

  \begin{icmlauthorlist}
    \icmlauthor{Sicheng Liu}{equal,yyy}
    \icmlauthor{Xunkai Li}{equal,yyy}
    \icmlauthor{Daohan Su}{yyy}
    \icmlauthor{Ru Zhang}{yyy}
    \icmlauthor{Hongchao Qin}{yyy}
    \icmlauthor{Ronghua Li}{yyy}
    \icmlauthor{Guoren Wang}{yyy}
  \end{icmlauthorlist}

  \icmlaffiliation{yyy}{Department of XXX, University of YYY, Location, Country}

  \icmlcorrespondingauthor{Ronghua Li}{lironghuabit@126.com}

  \icmlkeywords{Multimodal Attributed Graph, Graph Neural Network , Graph Foundation Model, Self-supervised Learning, ICML}

  \vskip 0.3in
]



\printAffiliationsAndNotice{\icmlEqualContribution}

\begin{abstract}
    Graph Foundation Models (GFMs) have achieved remarkable success in generalizing across diverse domains. However, they mainly focus on Text-Attributed Graphs (TAGs), leaving  Multimodal-Attributed Graphs (MAGs) largely untapped.
    Developing Multimodal Graph Foundation Models (MGFMs) allows for leveraging the rich multimodal  information in MAGs, and extends applicability to broader types of downstream tasks.
    While recent MGFMs integrate diverse modality information, our empirical investigation reveals two fundamental limitations of existing MGFMs:
    \ding{182} \textbf{they fail to explicitly model modality interaction}, essential for capturing intricate cross-modal semantics beyond simple aggregation, and \ding{183} \textbf{they exhibit sub-optimal modality alignment}, which is critical for bridging the significant semantic disparity between distinct modal spaces.
    
    To address these challenges, we propose \textbf{PLANET} (gra\underline{\textbf{P}}h topo\underline{\textbf{L}}ogy-aware mod\underline{\textbf{A}}lity i\underline{\textbf{N}}teraction and alignm\underline{\textbf{E}}n\underline{\textbf{T}}), 
    a novel framework employing a Divide-and-Conquer strategy to decouple modality interaction and alignment across distinct granularities.
    At the embedding granularity, \ding{182} \textbf{Embedding-wise Domain Gating (EDG)} performs local semantic enrichment by adaptively infusing topology-aware cross-modal context, achieving modality interaction.
    At the node granularity, \ding{183} \textbf{Node-wise Discretization Retrieval (NDR)} ensures global modality alignment by constructing a Discretized Semantic Representation Space (DSRS) to bridge modality gaps.
    Extensive experiments demonstrate that PLANET significantly outperforms state-of-the-art baselines across diverse graph-centric and multimodal generative tasks.
\end{abstract}

\section{Introduction}
\label{introduction}


In recent years, GFMs~\cite{xia2024opengraph,xia2024anygraph} have emerged as a transformative paradigm in graph representation learning, offering a unified encoding framework capable of generalizing across cross-domain datasets.
However, most existing GFMs are primarily designed for TAGs~\cite{he2024unigraph,wang2024gft} or focus on learning unified graph structures~\cite{yu2025samgpt,sun2025riemanngfm}, failing to extend effectively to MAGs~\cite{zhu2025mosaic,yan2025graph}. 
This limitation implies significant missed improvements. 
From a data perspective, MAGs enriched with diverse modalities offer significantly richer multimodal semantic information compared to TAGs.
From an application perspective, extending GFMs to MAGs significantly broadens the applicable downstream tasks such as modality retrieval task~\cite{qu2021dynamic} and graph generative task~\cite{yoon2023multimodal,fang2025graphgpt}.
To address this limitation, MGFMs~\cite{he2025unigraph2,fang2025graphgpt} have been introduced to integrate these heterogeneous signals within graph structures, achieve domain and modality alignment.
\begin{figure*}
    \centering
    \includegraphics[width=1.0\linewidth]{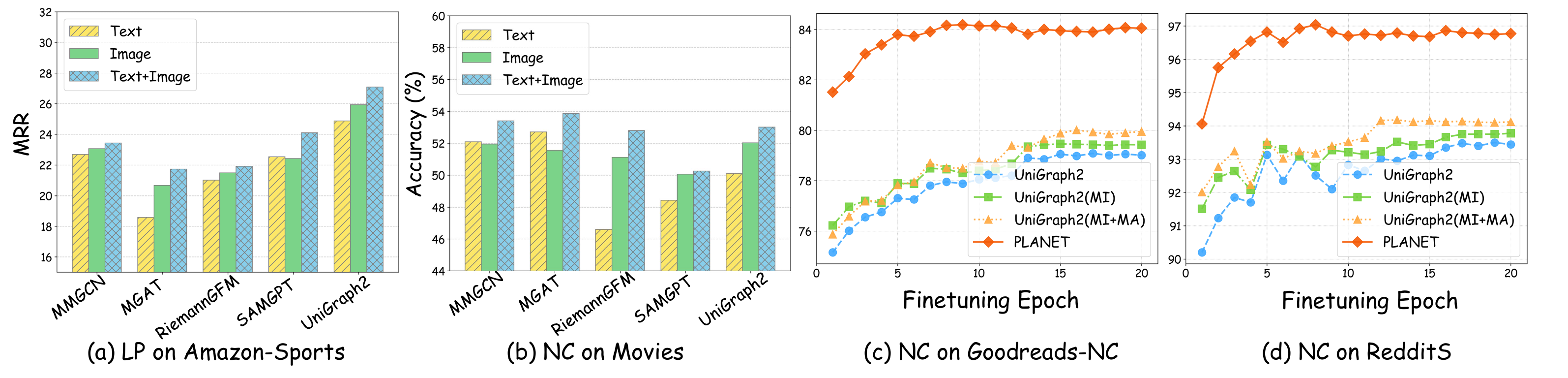}
    \vspace{-6mm}
    \caption{Empirical study results. (a)\&(b) Performance comparison across different modalities and architectures. (c)\&(d) Stepwise enhancement results.}
    \label{fig:fig1}
    \vspace{-6mm}
\end{figure*}

To validate the advantages of incorporating multimodal information and the superiority of MGFMs approaches, we conducted an empirical study shown in Fig.~\ref{fig:fig1}(a)\&(b).
Our findings yield two key conclusions:
\ding{182} The use of multiple modalities consistently improve model performance.
\ding{183} MGFMs significantly outperform GFMs.
UniGraph2~\cite{he2025unigraph2} consistently achieves state-of-the-art results, demonstrating the need for specialized architectures for MAGs. Please refer to Appendix~\ref{Appendix_emp_1} for detailed implementation.
Although MGFMs such as UniGraph2 have demonstrated a powerful capacity to learn from diverse multimodal graphs, 
they are fundamentally limited by their approach to modality interaction and alignment.
Specifically, UniGraph2 does not account for the significant semantic disparities between the latent spaces of different encoders and, crucially, lacks modality interaction, which is verified to be important in learning robust feature representations~\cite{zhu2025mosaic}.
Furthermore, UniGraph2 lacks explicit supervisory signals for modality alignment, rendering the alignment process inefficient and inadequately constrained.

To empirically substantiate these limitations of UniGraph2, we consider the general architecture of MGFMs, which typically consists of four core components:
\ding{182} Modality-specific encoding,
\ding{183} Modality Interaction (including Graph Neural Network (GNN) processing),
\ding{184} Modality Alignment,
and \ding{185} Modality Fusion. 
As detailed in our ablation studies (See Fig.~\ref{fig:fig1}(c)\&(d)), we involve a stepwise enhancement of the UniGraph2 architecture, progressively endowing it with capabilities for Modality Interaction (MI) and Modality Alignment (MA). 
The results demonstrate that each module provides a positive contribution to the model's overall performance.
This analysis forms the primary motivation for our work: 
to develop dedicated mechanisms for these two critical modules, tailored to address their challenges at distinct granularities.
Please refer to Appendix~\ref{Appendix_emp_2} for detailed implementation of the empirical study.
Motivated by the limitations mentioned above, 
in this work, we propose PLANET, a novel framework that utilizes a Divide-and-Conquer strategy to decouple the complexity inherent in MGFM design across two distinct granularities.
At the embedding granularity, we introduce the \ding{182} \textbf{EDG for Modality Interaction.} 
We formulate modality interaction as a process of local semantic enrichment.
This module employs a domain-specialized gating mechanism to adaptively extract and infuse topology-aware cross-modal context, enhancing the model's comprehension of intricate inter-modality relationships at the fine-grained embedding level.
At the node granularity, we propose the \ding{183} \textbf{NDR for Modality Alignment.} 
We view modality alignment as a global semantic consensus problem. 
By constructing a Discretized Semantic Representation Space (DSRS), this module retrieves and anchors heterogeneous signals into a unified space, explicitly enforcing robust alignment at the coarse-grained node level.


In summary, our \textbf{key contributions} in PLANET are:
(1) \textit{\underline{New Perspective.}} To the best of our knowledge, we are the first to systematically identify and address the critical shortcomings of existing MGFMs with respect to lack of modality interaction and alignment.
(2) \textit{\underline{Novel Approach.}} We propose PLANET, a novel framework that introduces a paradigm of topology-aware modality interaction and modality alignment,
offering a valuable reference for future designs in the MGFM domain.
(3) \textit{\underline{SOTA Performance.}} Extensive experiments demonstrate the superiority of PLANET across various graph-centric tasks and multimodal generative tasks.

\section{Preliminaries}
\label{sec:preliminaries}

\subsection{Notations and Problem Formulation}
\label{sec:2.1}
Consider a MAG denoted as $\mathcal{G} = (\mathcal{V}, \mathcal{E}, \mathcal{R})$, with $|\mathcal{V}| = N$ nodes and $|\mathcal{E}| = M$ edges. 
$\mathcal{R}$ denotes the collection of raw multimodal data associated with the nodes (e.g., raw text descriptions, images). 
Let $\mathcal{M} = \{m_1, m_2, \dots, m_{|\Omega|}\}$ be the set of available modalities, where $|\Omega|$ denotes the number of modalities. 
In this work, we operate under the assumption that each node $v \in \mathcal{V}$ possesses complete features across all modalities in $\mathcal{M}$, ensuring a comprehensive multimodal context for every entity.
The raw data is transformed into feature $\mathbf{X}^{(m)} \in \mathbb{R}^{N \times d_m}$ via $\phi_m$, where $\phi_m$ denotes the frozen modality-specific encoders for modality $m$ (e.g., ViT~\cite{dosovitskiy2020image} for images, BERT~\cite{devlin2019bert} for texts). 
Consequently, we obtain a feature-transformed graph $\mathcal{G}' = (\mathcal{V}, \mathcal{E}, \mathcal{X})$, where $\mathcal{X} = \{ \mathbf{X}^{(m)} \}_{m \in \mathcal{M}}$ serves as the input node features for the subsequent learning process.
 


Based on $\mathcal{G}'$, we aim to learn a unified embedding function $f_\theta$ following the standard paradigm: 
\uline{\textbf{Pre-training stage.}} We optimize $f_\theta$ using self-supervised objectives (e.g., reconstruction tasks and contrastive tasks). 
\uline{\textbf{Fine-tuning stage.}} The pre-trained $f_\theta$ is adapted with task-specific heads to support diverse downstream tasks.
\begin{figure*}[!t]
    \centering
    \includegraphics[width=1.0\linewidth]{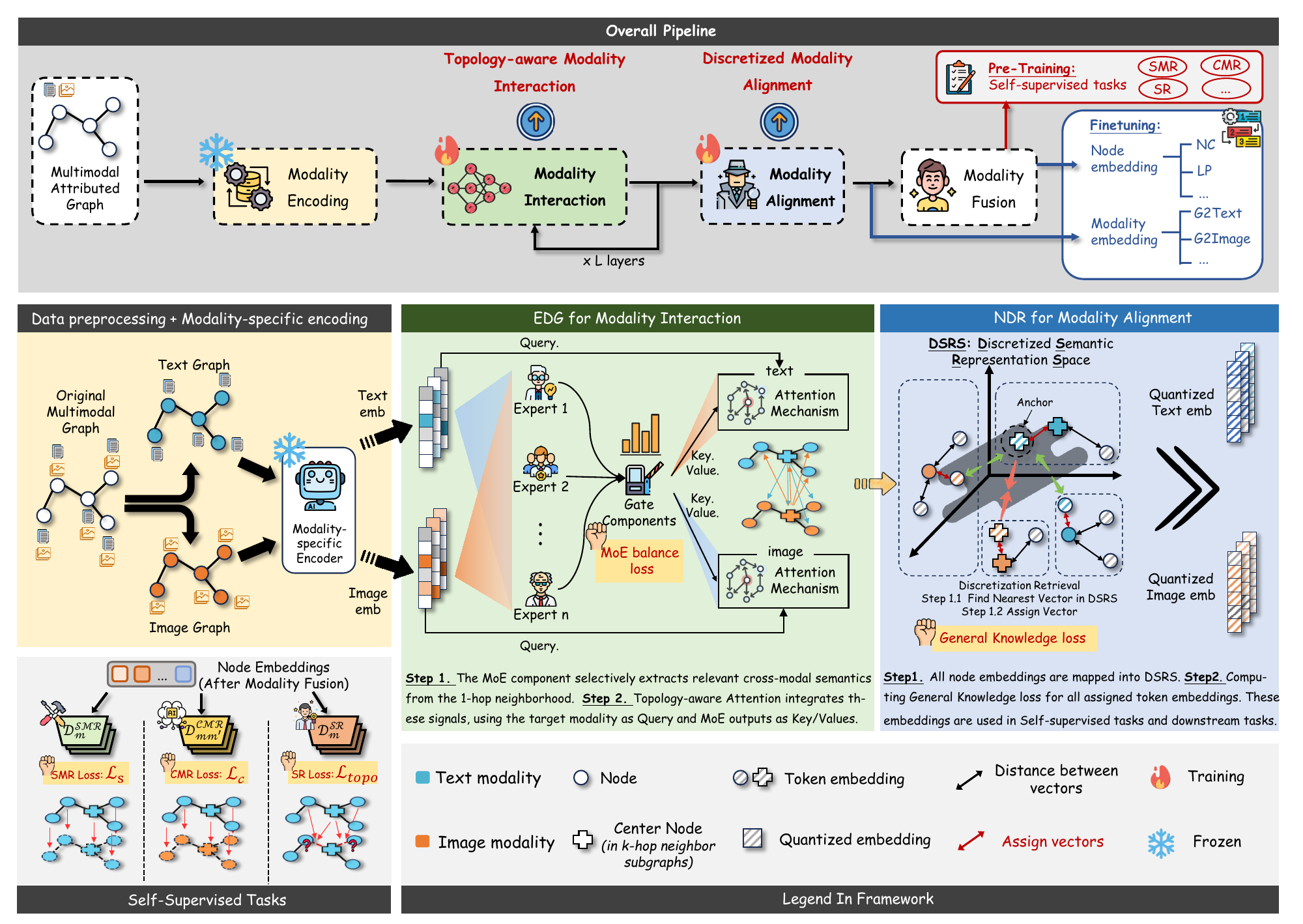}
    \vspace{-6mm}
    \caption{Overall architecture of the proposed method: PLANET.}
    \vspace{-3mm}
    \label{fig:framework}
\end{figure*}

\subsection{Multimodal Graph Learning}

Multimodal Graph Learning (MGL) is the machine learning technique on multimodal graphs~\cite{peng2024learning}. 
MGL are widely employed across various domains, including biology~\cite{gainza2020deciphering}, chemistry~\cite{guan2021regio}, knowledge graph~\cite{chen2022hybrid,zeng2023multi}, healthcare applications~\cite{zheng2022multi} and recommendation systems~\cite{wei2019mmgcn,tao2020mgat,yi2022multi,zhang2021mining}.
However, these existing models are inherently designed for specific tasks within specific domains (e.g., user-item link predictions), lacking the generalization capability to transfer effectively to other domains or downstream tasks.

\subsection{Graph Foundation Models}
Recently, Graph Foundation Models (GFMs) have gained significant attention for their robust generalization capabilities~\cite{liu2025graph}.
Some approaches focused on capturing universal structural patterns~\cite{chen2025towards,yu2025samgpt,sun2025riemanngfm}.
The majority of existing GFMs concentrate on TAGs, which focus on aligning graph structural representations with the semantic space~\cite{tang2024graphgpt,chen2024llaga,kong2024gofa,zhu2025graphclip,li2024zerog} and unifying diverse downstream tasks across domains to achieve transferability~\cite{liu2023ofa,huang2023prodigy}. However, these GFMs are not primarily designed for MAGs.

Unlike previous works, UniGraph2~\cite{he2025unigraph2} establishes a MGFM, but fails to effectively address the critical issues of modality interaction and alignment (see Sec.~\ref{introduction}). Although a new study GraphGPT-O~\cite{fang2025graphgpt} attempts to incorporate cross-modal interaction, it employs a non-topology-aware modality interaction mechanism within an LLM-based framework restricted to generative tasks, thereby suffering from prohibitive computational overhead and limited versatility.

\section{Methodology}
\label{sec:methodology}
\subsection{Overview}
\label{sec:overview}

We propose \textbf{PLANET}, 
a MGFM that uses a Divide-and-Conquer strategy to decouple modality interaction and modality alignment (Fig.~\ref{fig:framework}). 


\textbf{Modality Encoding.}
We use modality-specific encoders to transform raw multimodal data into feature vectors.
Following~\citet{hou2022graphmae}, we employ a modality masking strategy to these feature vectors (details in  Appendix~\ref{appendix: modality masking}), yielding the masked feature vectors $\tilde{\mathbf{X}}^{(m)}$.

For feature dimension alignment across heterogeneous modalities, we apply a set of modality-specific MLPs on $\tilde{\mathbf{X}}^{(m)}$:
$\mathbf{H}^{(0, m)} = \text{MLP}_m(\tilde{\mathbf{X}}^{(m)}).$
Subsequently, to preserve the inherent distribution of each modality, we process $\mathbf{H}^{(0, m)}$ through independent, modality-specific GNNs (e.g., Graph Transformer~\cite{dwivedi2020generalization}) to obtain the modality-specific embeddings $\mathbf{H}^{(spe,m)}$.


\textbf{Modality Interaction \& Modality Alignment.}
We first employ the EDG at the embedding granularity to infuse topology-aware cross-modal context (Sec.~\ref{sec:interaction}). 
Subsequently, the NDR operates at the node granularity, anchoring features into a DSRS for modality alignment. 
This process yields the aligned modality-level node representations $\mathbf{H}^{(cross, m)}$ (Sec.~\ref{sec:alignment}).

\textbf{Modality Fusion.}
To integrate modality specificity with interactive semantics, we first fuse modality embeddings: $\mathbf{H}^{(all,m)} = \mathbf{H}^{(spe,m)} \ || \ \mathbf{H}^{(cross,m)}$. 
Specifically, for graph-centric tasks (e.g., node classification), we further concatenate multimodal representations to obtain the final node embeddings: $\mathbf{h}_i = \|_{m \in \mathcal{M}} \mathbf{h}^{(all,m)}_i$.
The entire model is optimized via a joint self-supervised objective (Sec.~\ref{sec:training}).
\begin{figure}[!t]
    \centering
    \includegraphics[width=1.0\linewidth]{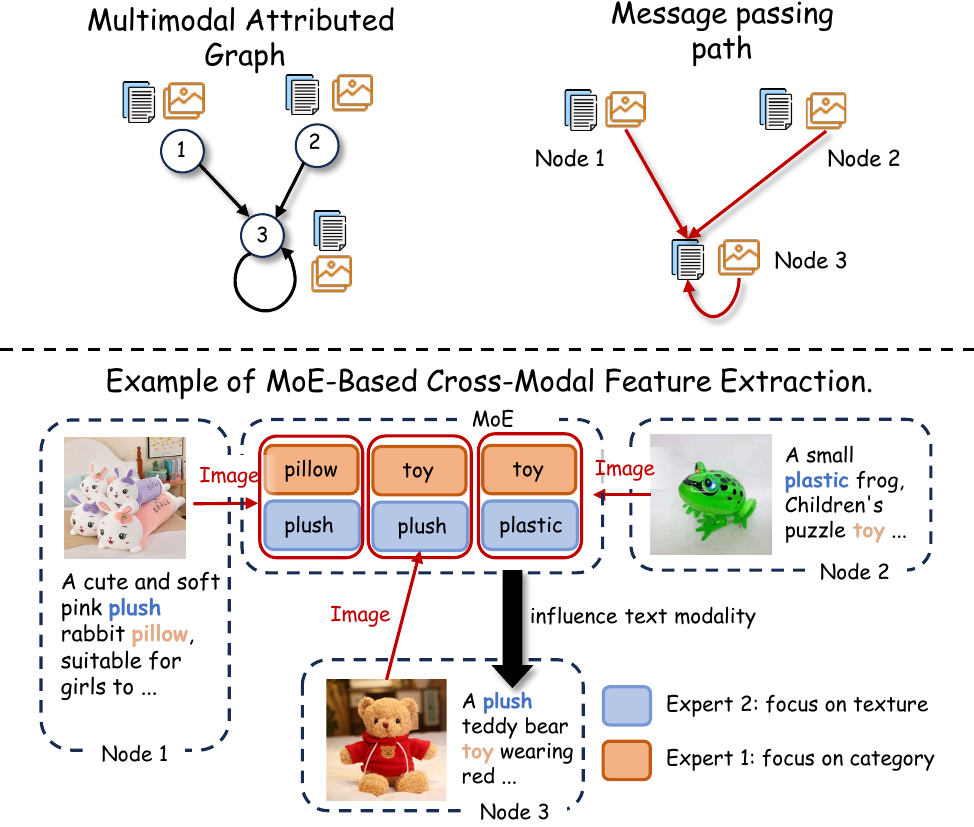}
    \vspace{-6mm}
    \caption{Illustration of expert-driven semantic extraction within the EDG module. Taking the text modality as the target example, we observe that the target text modality (in Node 3) correlates with neighboring images (in Node 1,2) via diverse attributes. Our MoE module is designed to capture these distinct semantic patterns through specialized experts, enabling the precise extraction of effective cross-modal mutual information.}
    \vspace{-8mm}
    \label{fig:moe_pic}
\end{figure}



\subsection{EDG for Modality Interaction}
\label{sec:interaction}

Following the Divide-and-Conquer strategy, the first challenge is to model modality interaction. We formulate this as a process of local semantic enrichment at the embedding granularity. Unlike previous works that aggregate modalities globally~\cite{he2025unigraph2}, EDG operates within the layer-wise propagation, allowing each node’s modality embedding to dynamically absorb complementary semantics from its neighbors before any global alignment is enforced.



\textbf{MoE-Based Cross-Modal Feature Extraction.}
Intuitively, in a MAG, the semantic representation of a node in one modality (e.g., text) is often correlated with the complementary modalities (e.g., image) of its neighbors and itself. 
For example, a textual description may correlate with the texture and category of objects in images (See Fig.~\ref{fig:moe_pic}).
Guided by this observation, we employ a Mixture-of-Experts (MoE) module to capture these diverse semantic patterns.
Formally, taking $\mathbf{H}^{(0,m)}$ as input, 
let $\mathbf{h}_i^{(\ell,m)}$ denote the output embedding of node $i$ for modality $m$ at the $l$-th layer of the EDG module. 


For a neighbor node $j$
and modality $m$.
The effective cross-modal signal $\mathbf{e}_j^{(\ell,m)}$ is computed as:
\vspace{-3mm}
\begin{equation}
\begin{split}
    &\mathbf{e}_j^{(\ell,m)} = \sum_{k=1}^{K} G\left(\mathbf{n}_j^{(\ell,m)}\right)_k \cdot E_k\left(\mathbf{n}_j^{(\ell,m)}\right), \\
    &\quad\quad \text{where} \; \mathbf{n}_j^{(\ell,m)} = \mathop{\Big\|}_{m' \neq m} \mathbf{h}_j^{(\ell-1,m')},
\end{split}
\end{equation}
and $E_k(\cdot)$ represents the $k$-th expert network (implemented as an MLP). 
Each expert specializes in discerning and extracting specific patterns of effective mutual information from the neighbor node set. $K$ denotes the total number of expert.
$G(\cdot)_k$ is the gating score indicating the relevance of the $k$-th expert, computed via a softmax gating network:
\begin{equation}
    G\left(\mathbf{n}_j^{(\ell,m)}\right)_k = \frac{\exp\left(\text{MLP}_g(\mathbf{n}_j^{(\ell,m)})_k\right)}{\sum_{k'=1}^{K} \exp\left(\text{MLP}_g(\mathbf{n}_j^{(\ell,m)})_{k'}\right)},
\end{equation}
This ensures that the model dynamically selects experts to capture the diverse semantic combinations of neighboring modalities.

\textbf{Topology-Aware Attention Mechanism.}
After mutual semantic information extraction, a Graph Transformer~\cite{dwivedi2020generalization} layer is used to perform topology-aware attention mechanism:
{
\setlength{\abovedisplayskip}{6pt} 
\setlength{\belowdisplayskip}{6pt} 
\small
\begin{equation}
    \mathbf{h}_i^{(\ell,m)} = \operatorname{GT}_{\ell} \left( q=\mathbf{h}_i^{(\ell-1,m)} ; \mathcal{K},\mathcal{V}=\{\mathbf{e}_j^{(\ell,m)}\}_{j\in\mathcal{N}_i}\right),
\end{equation}
}%
where $\mathcal{N}_i$ denotes the set of incoming neighbors for node $i$ (including node $i$ itself), $\operatorname{GT}_{l}(\cdot)$ represents the $l$-th layer of standard Graph Transformer.
Crucially, distinct from standard cross-modal attention which typically operates within isolated instances~\cite{radford2021learning}, our mechanism explicitly integrates graph structural information into the cross-modal interaction process by querying the complementary modalities of neighboring nodes.
See Appendix~\ref{appendix: EDG} for the detailed formulations of our attention mechanism.

\subsection{NDR for Modality Alignment}
\label{sec:alignment}
While the EDG module significantly enriches semantic representation by extracting and infusing cross-modal context, we argue that a robust MGFM requires a more explicit constraint to bridge the inherent semantic gap between modalities. 
We formulate modality alignment here as a global semantic consensus problem, to address this challenge, we introduce the NDR module at the node granularity.



\textbf{Discretization Retrieval.}
We define Discretized Semantic Representation Space (DSRS) as $\mathcal{S} = \{\mathbf{s}_1, \dots, \mathbf{s}_C\}$ containing $C$ learnable latent vectors, referred to as tokens. 
Let $\mathbf{h}^{(L,m)}_i$ denote the final output representation of node $i$ for modality $m$ from the EDG module. 
Each node embedding is mapped to its nearest DSRS token:
\begin{equation}
    \mathbf{h}^{(cross,m)}_i = \mathbf{s}_c, \; \text{where}\; c\! = \!\operatorname*{argmin}_j \|\mathbf{s}_j - \mathbf{h}^{(L,m)}_i \|_2,
\end{equation}
where $\|\cdot\|_2$ denotes the Euclidean distance.
This retrieval step forces features from different modalities to cluster around shared semantic space.

\textbf{Text-Anchored General Knowledge Alignment.}
Inspired by~\citet{liu2025multimodal}, language offers the most dense semantic information, we propose a \textit{General Knowledge Loss} that uses the \textit{Text} modality ($t$) as an anchor. 
We align every other modality $m$ to the text modality in the DSRS:
{
\begin{equation}
\small
\begin{aligned}
    \mathcal{L}_{gen}\!\!\!&=\!\!-\frac{1}{N(|\Omega|\!-\!1)}\!\sum_{m\neq t}\!\sum_{i=1}^{N}\!\!\left(\!\log\!\frac{z_{i,i}^{(t,m)}}{\sum_{j=1}^{N}\!z_{i,j}^{(t,m)}}\!+\!\log\!\frac{z_{i,i}^{(t,m)}}{\sum_{j=1}^{N}\!z_{j,i}^{(t,m)}}\!\right)\!, \\
    &\text{where} \quad z_{i,j}^{(t,m)} = \exp \left(\operatorname{sim}(\mathbf{h}_{i}^{(cross,t)},\mathbf{h}_{j}^{(cross,m)}) / \tau\right),
\end{aligned}
\end{equation}
}%
and $\tau$ is a temperature parameter. This efficiently pulls the quantized representations of the same node across modalities together while pushing distinct nodes apart. 

\textbf{VQ Objective.}
To align the latent distribution of modality features with the quantized semantic space of DSRS, we utilize the loss function~\cite{van2017neural}:
{
\begin{equation}
\label{eq: VQ loss}
\begin{aligned}
    \mathcal{L}_{VQ} = \frac{1}{N}\sum_{i=1}^{N} \sum_{m \in \mathcal{M}} \Big(  &\|sg[\mathbf{h}_{i}^{(cross,m)}]-\mathbf{h}_i^{(L,m)}\|_2^2 \\
    + &  \gamma \|\mathbf{h}_{i}^{(cross,m)}-  sg[\mathbf{h}_i^{(L,m)}]\|_2^2\Big),
\end{aligned}
\end{equation}
}

where $sg[\cdot]$ denotes the stop-gradient operator to handle the non-differentiable quantization, and $\gamma$ controls the commitment cost.

\subsection{Self-Supervised Training}
\label{sec:training}


\textbf{Feature Reconstruction.}
To ensure robust semantic understanding, we propose a dual reconstruction mechanism. We minimize the error for recovering the masked input itself ($\mathcal{L}_{self}$) to capture modality-specific semantics, while simultaneously enforcing cross-modal correlations by reconstructing features of complementary modalities:
{
\setlength{\abovedisplayskip}{3pt} 
\setlength{\belowdisplayskip}{3pt} 
\begin{equation}
\small
\begin{aligned}
    \mathcal{L}_{s} &= \frac{1}{|\Omega|N}\sum_{i=1}^N \sum_{m \in \mathcal{M}} \|\mathcal{D}_m^{SMR}(\mathbf{h}_{i}^{(all,m)})-\mathbf{x}_i^{(m)}\|_2^2, 
    \label{func:self_recon}
\end{aligned}
\end{equation}
}%
\begin{equation}
\small
\begin{aligned}
    \!\!\!\!\!\!\mathcal{L}_{c}\!\!\ &= \!\!\frac{\sum_{i=1}^N \sum_{m \neq m'} \|\mathcal{D}_{mm'}^{CMR}(\mathbf{h}_i^{(all,m)})\!-\!\!\mathbf{x}_i^{(m')}\|_2^2}{(|\Omega|^2 - |\Omega|)N},
    \label{func:cross_recon}
\end{aligned}
\end{equation}
where $\mathbf{x}_i^{(m)}$ is the original unmasked feature vector output by the modality encoder, $\mathcal{D}_m^{SMR}$ is the self modality reconstruction decoder (MLP), and $\mathcal{D}_{mm'}^{CMR}$ is the cross modality reconstruction decoder, both of which are implemented as MLPs. The total feature reconstruction loss is: $\mathcal{L}_{feat} = \mathcal{L}_{s} + \beta_{inter}\mathcal{L}_{c}$.

\textbf{Structural Reconstruction.}
To preserve topological information, we employ a link reconstruction task. We use $\mathcal{L}_{topo}$ to denote the loss function, which enforces higher similarity scores for connected edges compared to randomly sampled negative pairs (details in Appendix~\ref{appendix: loss}).

\textbf{Total Objective.}
The final training objective combines all losses:
\begin{equation}
\small
    \mathcal{L} = \beta_{1}\mathcal{L}_{feat} + \beta_{2}\mathcal{L}_{topo} + \beta_{3}\mathcal{L}_{gen} + \beta_{4}\mathcal{L}_{VQ} + \beta_{5}\mathcal{L}_{load},
\end{equation}
where $\beta$ terms are hyperparameters balancing the contribution of each component, $\mathcal{L}_{load}$ is load balancing loss introduced in Appendix~\ref{appendix: loss}.

\subsection{Theoretical Analysis}
\label{sec:theory}



\subsubsection{\small Why EDG Captures Synergistic Semantics?}


\begin{definition}
\textbf{Synergistic Features.}
\label{def:synergy}
Let $G^{(A)}$ and $G^{(B)}$ denote input graphs for modality $A$ and $B$. 
It can be decomposed into independent modality-specific unique features $\{U_A, U_B\}$ and synergistic features $\{S_A, S_B\}$. 
$\{S_A, S_B\}$ satisfies the condition of zero mutual information under independent views, i.e., $I(Y; S_A) \approx 0$ and $I(Y; S_B) \approx 0$, but positive interaction information under a joint view, i.e., $I(Y; S_A, S_B)>0$, where $Y$ represents the latent semantic information related to the data.
\end{definition}


\begin{theorem}
\textbf{Synergy Preservation via EDG.}
\label{thm:synergy}
Let $Z^*_{Vanilla}$ and $Z^*_{EDG}$ denote the optimal representations learned by a vanilla Multimodal Graph Encoder (e.g., MMGCN) and PLANET with the EDG module, respectively. Under the compression constraint of the Information Bottleneck~\cite{federici2020learning,wu2020graph}, provided that the trade-off parameter $\beta$ is sufficiently small, the information gap between the two representations satisfies:
\begin{equation}
\small
    I(Y; Z^*_{EDG}) - I(Y; Z^*_{Vanilla}) \ge I(Y; S_A, S_B \mid U_A, U_B) > 0.
\end{equation}
\end{theorem}
\vspace{-2mm}
Such a theorem demonstrates that vanilla encoders inevitably discard synergistic features $\{S_A, S_B\}$ as noise. In contrast, EDG explicitly captures them, yielding a strictly larger information gain (proofs are shown in Appendix~\ref{appendix: proof}). 

\begin{table*}[t]
\scriptsize 
\centering
\renewcommand\tabcolsep{2.5pt} 
\renewcommand{\arraystretch}{1.3}
\setlength{\aboverulesep}{0pt}
\setlength{\belowrulesep}{0pt}
\caption{Main results on node classification tasks. We report Accuracy and F1-Macro for each dataset. Best results are highlighted in \besttext{bold}, the second-best are marked with \secondbesttext{underline}. All baselines are first pre-trained and then fine-tuned.}
\vspace{-3mm}
\label{tab:nc_results_corrected}
\begin{tabular}{c | cc cc cc cc cc cc}
\toprule[1.1pt]
\multirow{2}{*}{\textbf{Method}} & \multicolumn{2}{c}{RedditS} & \multicolumn{2}{c}{Movies} & \multicolumn{2}{c}{Grocery} & \multicolumn{2}{c}{Toys} & \multicolumn{2}{c}{Ele-fashion} & \multicolumn{2}{c}{Goodreads-NC} \\
\cmidrule(lr){2-3} \cmidrule(lr){4-5} \cmidrule(lr){6-7} \cmidrule(lr){8-9} \cmidrule(lr){10-11} \cmidrule(lr){12-13}
& \textbf{Acc} & \textbf{F1-Macro} & \textbf{Acc} & \textbf{F1-Macro} & \textbf{Acc} & \textbf{F1-Macro} & \textbf{Acc} & \textbf{F1-Macro} & \textbf{Acc} & \textbf{F1-Macro} & \textbf{Acc} & \textbf{F1-Macro} \\
\midrule
GCN & 92.44{\tiny$\pm$0.62} & 87.34{\tiny$\pm$1.22} & 52.26{\tiny$\pm$0.74} & 42.35{\tiny$\pm$1.23} & 78.60{\tiny$\pm$0.41} & 64.90{\tiny$\pm$1.09} & 77.63{\tiny$\pm$0.72} & 75.54{\tiny$\pm$0.97} & 85.33{\tiny$\pm$0.06} & 68.01{\tiny$\pm$0.23} & 78.44{\tiny$\pm$0.06} & 68.12{\tiny$\pm$0.16} \\
\midrule
MMGCN & 90.27{\tiny$\pm$0.34} & 84.22{\tiny$\pm$0.73} & 53.41{\tiny$\pm$1.03} & 41.66{\tiny$\pm$2.03} & 82.56{\tiny$\pm$0.49} & 73.83{\tiny$\pm$0.93} & \secondbest{80.02{\tiny$\pm$0.64}} & 76.36{\tiny$\pm$1.23} & 86.59{\tiny$\pm$0.08} & 68.85{\tiny$\pm$0.35} & \secondbest{83.22{\tiny$\pm$0.10}} & 71.28{\tiny$\pm$0.22} \\
MGAT & 92.78{\tiny$\pm$0.50} & 87.27{\tiny$\pm$0.53} & \secondbest{53.87{\tiny$\pm$0.50}} & \secondbest{44.09{\tiny$\pm$1.60}} & \secondbest{83.74{\tiny$\pm$0.62}} & 74.77{\tiny$\pm$1.11} & 79.61{\tiny$\pm$0.74} & \secondbest{77.09{\tiny$\pm$0.87}} & 84.84{\tiny$\pm$0.08} & 69.62{\tiny$\pm$0.21} & 82.91{\tiny$\pm$0.04} & \secondbest{71.45{\tiny$\pm$0.11}} \\
\midrule
GRACE & 93.01{\tiny$\pm$0.53} & \secondbest{88.39{\tiny$\pm$1.12}} & 48.09{\tiny$\pm$0.97} & 37.18{\tiny$\pm$1.33} & 70.83{\tiny$\pm$0.81} & 60.69{\tiny$\pm$1.05} & 72.82{\tiny$\pm$0.66} & 69.09{\tiny$\pm$0.63} & 83.58{\tiny$\pm$0.11} & 70.09{\tiny$\pm$0.47} & 74.96{\tiny$\pm$0.06} & 70.09{\tiny$\pm$0.11} \\
GraphMAE2 & 92.81{\tiny$\pm$0.44} & 87.93{\tiny$\pm$0.37} & 50.08{\tiny$\pm$0.77} & 38.68{\tiny$\pm$1.63} & 76.24{\tiny$\pm$0.60} & 66.74{\tiny$\pm$1.33} & 75.11{\tiny$\pm$0.52} & 71.80{\tiny$\pm$0.50} & 83.32{\tiny$\pm$0.31} & 65.92{\tiny$\pm$0.59} & 74.15{\tiny$\pm$0.22} & 69.20{\tiny$\pm$0.28} \\
\midrule
RiemannGFM & 91.63{\tiny$\pm$0.45} & 85.20{\tiny$\pm$1.13} & 52.80{\tiny$\pm$0.43} & 40.74{\tiny$\pm$1.25} & 82.62{\tiny$\pm$0.46} & \secondbest{74.90{\tiny$\pm$1.55}} & 77.85{\tiny$\pm$0.45} & 74.84{\tiny$\pm$0.60} & 87.07{\tiny$\pm$0.20} & \secondbest{70.45{\tiny$\pm$1.32}} & 78.13{\tiny$\pm$0.17} & 70.73{\tiny$\pm$0.24} \\
GFT & 93.02{\tiny$\pm$0.37} & 87.00{\tiny$\pm$2.03} & 51.33{\tiny$\pm$0.67} & 28.14{\tiny$\pm$1.82} & 76.80{\tiny$\pm$2.22} & 59.11{\tiny$\pm$3.02} & 79.52{\tiny$\pm$0.58} & 76.00{\tiny$\pm$0.92} & \secondbest{87.14{\tiny$\pm$0.22}} & 70.33{\tiny$\pm$1.24} & 75.93{\tiny$\pm$0.41} & 66.18{\tiny$\pm$0.30} \\
SAMGPT & 93.11{\tiny$\pm$0.19} & 87.12{\tiny$\pm$0.64} & 50.25{\tiny$\pm$0.28} & 34.21{\tiny$\pm$1.37} & 76.41{\tiny$\pm$0.54} & 63.40{\tiny$\pm$0.88}  & 73.81{\tiny$\pm$0.41} & 67.12{\tiny$\pm$0.73} & 83.81{\tiny$\pm$0.11} & 69.73{\tiny$\pm$0.39} & 74.29{\tiny$\pm$0.11} & 66.57{\tiny$\pm$0.29} \\
\midrule
UniGraph2 & \secondbest{93.65{\tiny$\pm$0.17}} & 87.91{\tiny$\pm$0.68} & 53.02{\tiny$\pm$0.53} & 43.43{\tiny$\pm$1.86} & 82.10{\tiny$\pm$0.37} & 73.93{\tiny$\pm$1.37} & 79.00{\tiny$\pm$0.59} & 76.02{\tiny$\pm$0.78} & 87.06{\tiny$\pm$0.18} & 69.80{\tiny$\pm$0.82} & 79.06{\tiny$\pm$0.27} & 68.74{\tiny$\pm$0.36} \\
PLANET & \best{96.62{\tiny$\pm$0.22}} & \best{92.44{\tiny$\pm$0.43}} & \best{57.06{\tiny$\pm$0.61}} & \best{47.49{\tiny$\pm$1.23}} & \best{85.16{\tiny$\pm$0.88}} & \best{77.23{\tiny$\pm$0.86}} & \best{81.22{\tiny$\pm$0.50}} & \best{77.55{\tiny$\pm$0.75}} & \best{87.37{\tiny$\pm$0.12}} & \best{70.74{\tiny$\pm$0.78}} & \best{84.16{\tiny$\pm$0.07}} & \best{74.43{\tiny$\pm$0.18}} \\
\midrule
\textbf{Rel-Improv.} & \textcolor{upcolor}{$\uparrow$ 3.17\%} & \textcolor{upcolor}{$\uparrow$ 4.58\%} & \textcolor{upcolor}{$\uparrow$ 5.92\%} & \textcolor{upcolor}{$\uparrow$ 7.71\%} & \textcolor{upcolor}{$\uparrow$ 1.70\%} & \textcolor{upcolor}{$\uparrow$ 3.11\%} & \textcolor{upcolor}{$\uparrow$ 1.50\%} & \textcolor{upcolor}{$\uparrow$ 0.60\%} & \textcolor{upcolor}{$\uparrow$ 0.26\%} & \textcolor{upcolor}{$\uparrow$ 0.41\%} & \textcolor{upcolor}{$\uparrow$ 1.13\%} & \textcolor{upcolor}{$\uparrow$ 4.17\%} \\
\bottomrule[1.1pt]
\end{tabular}
\vspace{-3mm}
\end{table*}

\begin{table*}[t]
\small
\centering
\renewcommand\tabcolsep{4.5pt} 
\renewcommand{\arraystretch}{1.1} 
\setlength{\aboverulesep}{0pt}
\setlength{\belowrulesep}{0pt}
\caption{Link prediction and few-shot link classification results. We report MRR for link prediction, accuracy for few-shot link classification tasks.}
\vspace{-3mm}
\label{tab:few_shot_lp_results}
\begin{tabular}{cc|ccc ccc ccc}
\toprule[1.1pt]
& \multirow{3}{*}{\textbf{Method}} & \multicolumn{3}{c}{Link Prediction} & \multicolumn{6}{c}{Few-shot Link Classification}\\
\cmidrule(lr){3-5} \cmidrule{6-11}
& & \multirow{2}{*}{\makecell[c]{Amazon \\-Sports}} & \multirow{2}{*}{\makecell[c]{Amazon \\-Cloth}} & \multirow{2}{*}{\makecell[c]{Goodreads \\-LP}} & \multicolumn{3}{c}{Amazon-Sports-2Way} & \multicolumn{3}{c}{Amazon-Cloth-2Way} \\
\cmidrule(lr){6-8} \cmidrule(lr){9-11}
& & & & & \textbf{10-shot} & \textbf{5-shot} & \textbf{3-shot} & \textbf{10-shot} & \textbf{5-shot} & \textbf{3-shot} \\
\midrule
& MMGCN & 23.44{\tiny$\pm$0.43} & 17.74{\tiny$\pm$0.38} & 20.73{\tiny$\pm$0.48} & 56.66{\tiny$\pm$1.60} & 53.70{\tiny$\pm$1.20} & 55.86{\tiny$\pm$2.96} & 67.27{\tiny$\pm$4.20} & 68.61{\tiny$\pm$5.38} & 64.36{\tiny$\pm$4.07}\\
& MGAT & 21.74{\tiny$\pm$0.96} & 15.47{\tiny$\pm$0.32} & 21.82{\tiny$\pm$0.53} & 57.92{\tiny$\pm$1.80} & 56.55{\tiny$\pm$2.69} & 54.92{\tiny$\pm$2.40} & 68.66{\tiny$\pm$2.94} & 70.34{\tiny$\pm$3.36} & 67.05{\tiny$\pm$1.45} \\
\midrule
& GRACE & 25.31{\tiny$\pm$0.16} & 18.27{\tiny$\pm$0.15} & 19.30{\tiny$\pm$0.27} & 58.50{\tiny$\pm$1.43} & 57.94{\tiny$\pm$2.58} & 56.42{\tiny$\pm$1.36} & 64.96{\tiny$\pm$1.86} & 64.94{\tiny$\pm$1.09} & 62.36{\tiny$\pm$2.75} \\
& GraphMAE2 & 24.54{\tiny$\pm$0.30} & 18.69{\tiny$\pm$0.21} & 19.99{\tiny$\pm$0.19} & 56.05{\tiny$\pm$1.70} & 54.36{\tiny$\pm$1.65} & 53.28{\tiny$\pm$3.09} & 62.33{\tiny$\pm$1.78} & 61.39{\tiny$\pm$1.71} & 60.95{\tiny$\pm$1.61} \\
\midrule
\multirow{3}{*}{\scriptsize \rotatebox{90}{GFM}} & RiemannGFM & 21.92{\tiny$\pm$0.51} & 19.20{\tiny$\pm$0.44} & 22.03{\tiny$\pm$0.67} & 53.68{\tiny$\pm$1.24} & 53.73{\tiny$\pm$2.03} & 54.07{\tiny$\pm$1.46} & 66.20{\tiny$\pm$5.94} & 62.66{\tiny$\pm$1.75} & 63.45{\tiny$\pm$4.07}\\
& GFT & 22.04{\tiny$\pm$0.62} & 17.63{\tiny$\pm$0.59} & 20.16{\tiny$\pm$1.21} & 55.73{\tiny$\pm$3.36} & 56.86{\tiny$\pm$2.30} & 56.75{\tiny$\pm$2.63} & 65.37{\tiny$\pm$4.03} & 66.66{\tiny$\pm$2.10} & 63.70{\tiny$\pm$2.84} \\
& SAMGPT & 24.09{\tiny$\pm$0.22} & 16.41{\tiny$\pm$0.20} & \secondbest{24.91{\tiny$\pm$0.38}} & 60.48{\tiny$\pm$4.25} & 59.00{\tiny$\pm$3.56} & 59.41{\tiny$\pm$3.62} & \secondbest{74.25{\tiny$\pm$1.75}} & \secondbest{74.20{\tiny$\pm$1.60}} & 72.61{\tiny$\pm$1.90} \\
\midrule
\multirow{2}{*}{\scriptsize \rotatebox{90}{MGFM}} & UniGraph2 & \secondbest{27.09{\tiny$\pm$0.13}} & \secondbest{19.31{\tiny$\pm$0.35}} & 19.44{\tiny$\pm$0.19} & \secondbest{65.08{\tiny$\pm$3.17}} & \secondbest{64.27{\tiny$\pm$2.74}} & \secondbest{60.83{\tiny$\pm$3.90}} & 73.77{\tiny$\pm$1.91} & 71.28{\tiny$\pm$4.20} & \secondbest{73.44{\tiny$\pm$1.57}} \\
& PLANET & \best{27.51{\tiny$\pm$0.14}}  & \best{20.25{\tiny$\pm$0.27}} & \best{27.62{\tiny$\pm$0.25}} & \best{67.84{\tiny$\pm$1.38}} & \best{64.36{\tiny$\pm$3.05}} & \best{62.89{\tiny$\pm$3.99}} & \best{75.44{\tiny$\pm$2.84}} & \best{75.22{\tiny$\pm$1.20}} & \best{74.03{\tiny$\pm$4.12}} \\
\midrule
& \textbf{Rel-Improv.} & \textcolor{upcolor}{$\uparrow$ 1.55\%} & \textcolor{upcolor}{$\uparrow$ 4.87\%} & \textcolor{upcolor}{$\uparrow$ 10.87\%} & \textcolor{upcolor}{$\uparrow$ 4.24\%} & \textcolor{upcolor}{$\uparrow$ 0.14\%} & \textcolor{upcolor}{$\uparrow$ 3.39\%} & \textcolor{upcolor}{$\uparrow$ 1.60\%} & \textcolor{upcolor}{$\uparrow$ 1.37\%} & \textcolor{upcolor}{$\uparrow$0.80\%} \\
\bottomrule[1.1pt]
\end{tabular}
\vspace{-4mm}
\end{table*}

\subsubsection{\small How NDR Enhances Alignment Efficiency?}
\begin{definition}
\textbf{Push-forward Measure.}
\label{def:proxy}
Let $\hat{\mu}_m = \frac{1}{N} \sum_{i=1}^N \delta_{x_i^{(m)}}$ be the empirical measure of modality $m$, where $\delta$ is the Dirac measure, $N$ is the number of samples. 
Given a DSRS $\mathcal{S}$ and a quantization function $Q(x) = \arg\min_{e \in \mathcal{S}} \|x - e\|_2$, we define the \textbf{Push-forward Measure} of modality $m$ as $\hat{\nu}_m = Q_{\#} \hat{\mu}_m = \frac{1}{N} \sum_{i=1}^N \delta_{Q(x_i^{(m)})}.$
\end{definition}

\begin{theorem}
\textbf{Efficient Alignment via NDR.}
\label{thm:alignment}
Assume the feature space is bounded. The alignment error between the empirical distribution of modality $m$ ($\hat{\mu}_m$) and the anchor text modality $t$ ($\hat{\mu}_t$) is bounded by:
{
\setlength{\abovedisplayskip}{3pt} 
\setlength{\belowdisplayskip}{3pt} 
\begin{equation}
\small
\begin{aligned}
\label{eq: theorem2}
    W_1(\hat{\mu}_m, \hat{\mu}_t)\!\! &\le \!\mathbb{E}_{x \sim \hat{\mu}_m} \|x - Q(x)\|_2 + \mathbb{E}_{z \sim \hat{\mu}_t}\|z - Q(z)\|_2 \!+ \\
    & W_1(\nu_m^{*},\nu_t^{*}) + O\left(\frac{C}{\sqrt{N}}\right),
\end{aligned}
\end{equation}
}%
where $W_1(\cdot, \cdot)$ denotes the 1-Wasserstein distance between distributions, $C$ represents the DSRS size, and $W_1(\nu_m^{*},\nu_t^{*})$ represents the intrinsic bias between modalities.
\end{theorem}
This proves that projecting features into DSRS accelerates the alignment convergence rate from $O(N^{-1/d})$ in continuous spaces to $O(N^{-1/2})$. Additionally, minimizing $\mathcal{L}_{VQ}$ reduces the quantization error terms in Eq.~\eqref{eq: theorem2}, explicitly tightening the upper bound to ensure robust alignment.

\section{Experiments}

To validate the superiority of PLANET, we raise several questions:
    \textbf{Q1}: Does PLANET consistently outperform SOTA baselines across standard graph-centric tasks (i.e., node classification and link prediction) under both supervised learning and few-shot learning scenarios?
    \textbf{Q2}: How do the proposed \textbf{EDG} and \textbf{NDR} modules contribute to achieving topology-aware modality interaction and robust modality alignment, respectively?
    \textbf{Q3}: Can PLANET effectively support downstream generative tasks,  thereby demonstrating the robust generalization capabilities and versatile applicability of a foundation model?
    \textbf{Q4}: How does PLANET fare in terms of computational efficiency compared to existing GFMs?
    The implementation details are introduced in Appendix~\ref{experiment settings}.

\begin{table*}[t]
\small 
\centering
\renewcommand\tabcolsep{3.9pt} 
\renewcommand{\arraystretch}{1.1} 
\setlength{\aboverulesep}{0pt}
\setlength{\belowrulesep}{0pt}
\caption{Few-shot node classification results. We report accuracy for node classification tasks.}
\vspace{-2mm}
\label{tab:few_shot_results}
\begin{tabular}{cc| ccc ccc ccc}
\toprule[1.1pt]
& \multirow{2}{*}{\textbf{Method}} & \multicolumn{3}{c}{Grocery-5way} & \multicolumn{3}{c}{Ele-fashion-5way} & \multicolumn{3}{c}{Goodreads-NC-5way} \\
\cmidrule(lr){3-5} \cmidrule(lr){6-8} \cmidrule(l){9-11}
 & & \textbf{10-shot} & \textbf{5-shot} & \textbf{3-shot} & \textbf{10-shot} & \textbf{5-shot} & \textbf{3-shot} & \textbf{10-shot} & \textbf{5-shot} & \textbf{3-shot} \\
\midrule
& MMGCN & 53.73{\tiny$\pm$3.51} & 50.30{\tiny$\pm$3.13} & 48.13{\tiny$\pm$2.65} & 60.87{\tiny$\pm$3.08} & 57.05{\tiny$\pm$2.42} & 54.37{\tiny$\pm$2.99} & 56.42{\tiny$\pm$2.84} & 54.10{\tiny$\pm$2.90} & 52.93{\tiny$\pm$2.94} \\
& MGAT & 55.53{\tiny$\pm$2.98} & 54.07{\tiny$\pm$3.18} & 50.83{\tiny$\pm$2.88} & 62.82{\tiny$\pm$2.18} & 61.10{\tiny$\pm$2.20} & \secondbest{60.38{\tiny$\pm$2.57}} & 58.22{\tiny$\pm$3.40} & 57.05{\tiny$\pm$2.35} & 53.15{\tiny$\pm$3.09} \\
\midrule
& GRACE & 62.22{\tiny$\pm$2.89} & 59.22{\tiny$\pm$3.50} & 57.28{\tiny$\pm$4.91} & \secondbest{64.72{\tiny$\pm$3.04}} & 59.90{\tiny$\pm$2.73} & 55.49{\tiny$\pm$4.10} & 59.20{\tiny$\pm$2.00} & 61.53{\tiny$\pm$3.14} & 56.84{\tiny$\pm$4.65} \\
& GraphMAE2 & 58.00{\tiny$\pm$4.26} & 54.60{\tiny$\pm$3.83} & 51.90{\tiny$\pm$5.76} & 63.65{\tiny$\pm$3.48} & 60.00{\tiny$\pm$3.53} & 57.35{\tiny$\pm$3.05} & 49.83{\tiny$\pm$2.60} & 46.30{\tiny$\pm$2.95} & 43.20{\tiny$\pm$2.70} \\
\midrule
\multirow{3}{*}{\scriptsize \rotatebox{90}{GFM}} & RiemannGFM & \secondbest{67.14{\tiny$\pm$2.02}} & \secondbest{66.23{\tiny$\pm$2.73}} & \secondbest{63.63{\tiny$\pm$2.48}} & 61.73{\tiny$\pm$2.90} & 60.24{\tiny$\pm$3.55} & 59.13{\tiny$\pm$3.71} & 60.54{\tiny$\pm$3.92} & 57.17{\tiny$\pm$3.83} & 54.29{\tiny$\pm$4.10} \\
& GFT & 63.12{\tiny$\pm$4.07} & 64.60{\tiny$\pm$3.54} & 61.45{\tiny$\pm$2.22} & 62.28{\tiny$\pm$2.62} & \secondbest{61.38{\tiny$\pm$3.41}} & 59.08{\tiny$\pm$3.95} & 51.62{\tiny$\pm$4.73} & 50.95{\tiny$\pm$4.35} & 48.70{\tiny$\pm$4.96} \\
& SAMGPT & 66.47{\tiny$\pm$10.67} & 62.33{\tiny$\pm$10.49} & 52.73{\tiny$\pm$12.69} & 61.27{\tiny$\pm$10.82} & 61.13{\tiny$\pm$9.32} & 54.20{\tiny$\pm$10.78} & 51.80{\tiny$\pm$8.65} & 47.00{\tiny$\pm$8.35} & 42.53{\tiny$\pm$6.79} \\
\midrule
\multirow{2}{*}{\scriptsize \rotatebox{90}{MGFM}} & UniGraph2 & 66.27{\tiny$\pm$3.11} & 61.25{\tiny$\pm$3.09} & 60.05{\tiny$\pm$4.13} & 60.38{\tiny$\pm$2.08} & 58.73{\tiny$\pm$2.95} & 53.98{\tiny$\pm$4.07} & \secondbest{63.80{\tiny$\pm$4.02}} & \secondbest{61.55{\tiny$\pm$3.97}} & \secondbest{59.67{\tiny$\pm$4.23}} \\
& PLANET & \best{81.93{\tiny$\pm$3.48}} & \best{79.88{\tiny$\pm$3.27}} & \best{77.85{\tiny$\pm$4.06}} & \best{74.85{\tiny$\pm$3.73}} & \best{72.97{\tiny$\pm$3.55}} & \best{70.50{\tiny$\pm$4.37}} & \best{69.18{\tiny$\pm$3.90}} & \best{67.59{\tiny$\pm$4.68}} & \best{63.62{\tiny$\pm$4.11}} \\
\midrule
& \textbf{Rel-Improv.} & \textcolor{upcolor}{$\uparrow$ 22.03\%} & \textcolor{upcolor}{$\uparrow$ 20.61\%} & \textcolor{upcolor}{$\uparrow$ 22.35\%} & \textcolor{upcolor}{$\uparrow$ 15.65\%} & \textcolor{upcolor}{$\uparrow$ 18.88\%} & \textcolor{upcolor}{$\uparrow$ 16.76\%} & \textcolor{upcolor}{$\uparrow$ 8.43\%} & \textcolor{upcolor}{$\uparrow$ 9.81\%} & \textcolor{upcolor}{$\uparrow$ 6.62\%} \\
\bottomrule[1.1pt]
\end{tabular}
\vspace{-4mm}
\end{table*}

\subsection{Graph-Centric Tasks}
\textbf{Experimental Settings.}
To answer \textbf{Q1}, we conduct extensive evaluations on node classification and link prediction tasks under both supervised learning and few-shot learning scenarios. 
We compare PLANET with 9 strong baselines, which can be categorized into five distinct groups:
    (1) \textbf{Vanilla GNNs}: GCN~\cite{kipf2016gcn}.
    (2) \textbf{Multimodal Graph Models}: Including MMGCN~\cite{wei2019mmgcn} and MGAT~\cite{tao2020mgat}.
    (3) \textbf{Self-supervised Graph Learning Models}: Comprising contrastive learning method, GRACE~\cite{zhu2020deep} and generative learning method, GraphMAE2~\cite{hou2023graphmae2}.
    (4) \textbf{Graph Foundation Models}: Including SAMGPT~\cite{yu2025samgpt} and RiemannGFM~\cite{sun2025riemanngfm}, which focus on learning unified graph structures, and GFT~\cite{wang2024gft}, which targets TAGs.
    (5) \textbf{Multimodal Graph Foundation Models}: Specifically UniGraph2~\cite{he2025unigraph2}, a novel baseline explicitly designed for processing MAGs.
Please refer to Appendix~\ref{experiment settings} for detailed implementation of baselines.

\textbf{Supervised Learning.}
Table~\ref{tab:nc_results_corrected} and~\ref{tab:few_shot_lp_results} show the results of node classification and link prediction in supervised learning.
The results indicate that PLANET consistently outperforms baselines across all datasets.
We attribute this superior performance to the synergistic collaboration between the EDG and NDR modules operating at distinct granularities. 
By jointly facilitating efficient topology-aware modality interaction and alignment, PLANET ensures the generation of high-quality node embeddings.

\begin{figure*}
    \centering
    \includegraphics[width=1.0\linewidth]{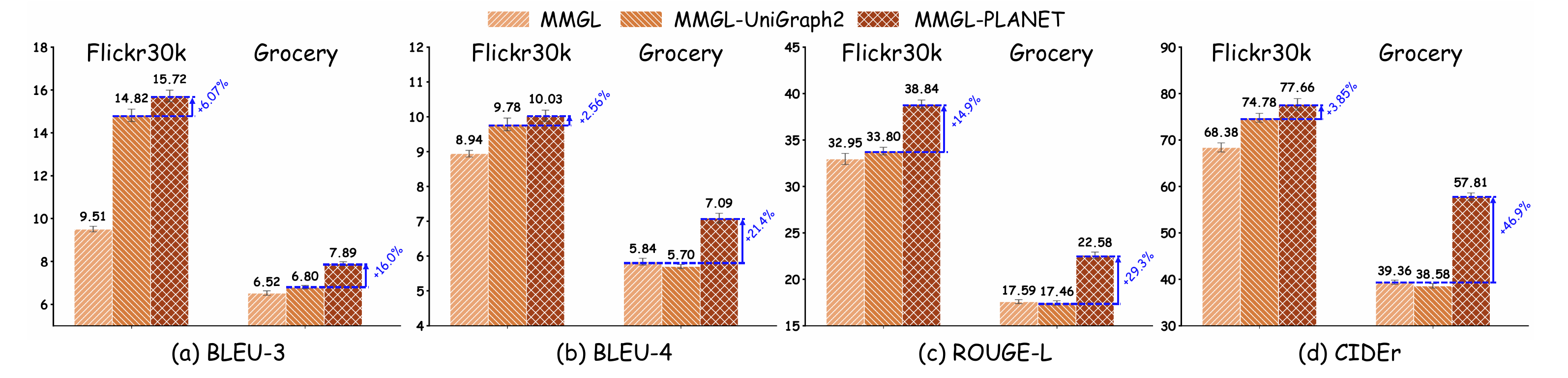}
    \caption{G2Text generation results. We report BLEU-3, BLEU-4, ROUGE-L, and CIDEr on the Flickr30k and Grocery datasets.}
    \vspace{-3mm}
    \label{fig:g2text}
\end{figure*}

\begin{figure*}
    \centering
    \includegraphics[width=1.0\linewidth]{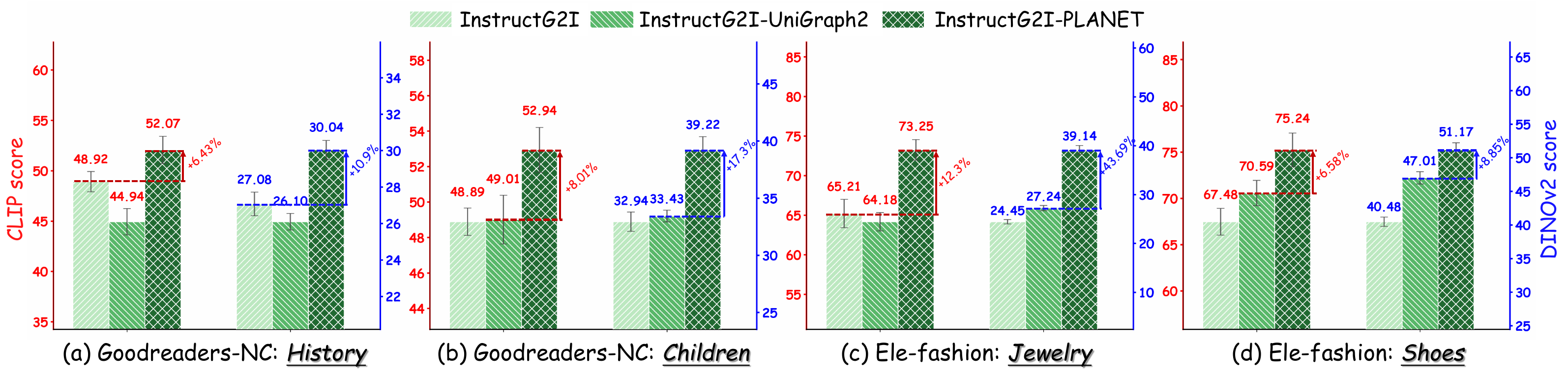}
    \caption{G2Image generation results. We report CLIP scores in \textcolor{red}{red} and DINOv2 scores in \textcolor{blue}{blue} across four categories selected from the Goodreads-NC and Ele-fashion datasets.}
    \vspace{-6mm}
    \label{fig:g2image}
\end{figure*}

\textbf{Few-Shot Learning.}
Tables~\ref{tab:few_shot_lp_results} and \ref{tab:few_shot_results} present the results for few-shot node classification and link prediction, respectively.
In these challenging low-resource scenarios, PLANET continues to outperform baselines across all datasets.
Most notably, in few-shot node classification tasks, PLANET achieves remarkable performance gains (15.68\%).
These results underscore the model's robust learning and generalization capabilities under conditions of severe data scarcity, 
providing evidence that our pre-training paradigm effectively enables the model to acquire comprehensive prior knowledge, which can be efficiently transferred to downstream tasks with minimal supervision.

\subsection{Ablation Study}
\label{sec: Ablation Study}


\begin{table}[t]
\centering
\small
\renewcommand{\arraystretch}{1.2} 
\renewcommand\tabcolsep{1.2pt}
\caption{Ablation studies on PLANET key components.}
\vskip -0.1in
\label{tab:ablation study}
\begin{tabular}{lcccc}
\toprule[1.1pt]
    & Toys & RedditS & Ele-fashion  & Amazon-Sports \\
\midrule
    \multicolumn{5}{l}{\textit{EDG for Modality Interaction}} \\ 
    w/o. MoE & 77.77{\tiny$\pm$0.64} & 95.19{\tiny$\pm$0.30} & 85.33{\tiny$\pm$0.27} & 25.92{\tiny$\pm$0.16}\\
    w/o. MoE+AM & 76.33{\tiny$\pm$0.54} & 93.01{\tiny$\pm$0.16} & 84.52{\tiny$\pm$0.14} & 24.76{\tiny$\pm$0.13}\\
    \midrule 
    \multicolumn{5}{l}{\textit{NDR for Modality Alignment}} \\ 
    w/o. DSRS& 78.53{\tiny$\pm$0.80} & 95.12{\tiny$\pm$0.33} & 85.93{\tiny$\pm$0.18} & 26.19{\tiny$\pm$0.22}\\
    w/o. $\mathcal{L}_{gen}$ & 77.85{\tiny$\pm$0.73} & 94.97{\tiny$\pm$0.31} & 84.71{\tiny$\pm$0.22} & 25.38{\tiny$\pm$0.19}\\
    \midrule 
    PLANET & \textbf{81.22{\tiny$\pm$0.50}} & \textbf{96.62{\tiny$\pm$0.22}} & \textbf{87.37{\tiny$\pm$0.12}} & \textbf{27.51{\tiny$\pm$0.14}}\\
\bottomrule[1.1pt]
\end{tabular}
\vspace{-6mm}
\end{table}

To address \textbf{Q2}, we conduct rigorous ablation studies removing core components of PLANET, as shown in Table~\ref{tab:ablation study}. 

\ding{182} \textit{\textbf{w/o. MoE Component.}} 
We remove the MoE mechanism. 
Here, neighboring multimodal features are fed directly into the attention mechanism.
The result confirms that the MoE is essential for extracting effective cross-modal signals before modality interaction.
\ding{183} \textit{\textbf{w/o. MoE + Attention Mechanism.}} We investigate the impact of topology-aware modality interaction by replacing the entire EDG module with standard, independent Graph Transformers for each modality.
The significant performance degradation validates the effectiveness of explicitly modeling the topology-aware interplay between modalities.
\ding{184} \textit{\textbf{w/o. DSRS.}} We remove the DSRS and the associated Discretization Retrieval, directly applying the General Knowledge Loss ($\mathcal{L}_{gen}$) to the continuous embeddings $\mathbf{h}_i^{(L,m)}$. The resulting performance drop confirms its critical role in modality alignment by projecting distinct modalities into a unified semantic space.
\ding{185} \textit{\textbf{w/o. General Knowledge Loss.}} We retain the DSRS structure but disable the text-anchored General Knowledge Loss ($\mathcal{L}_{gen}$). 
Results show that without this signal, the model fails to effectively bridge the semantic gap between the quantized tokens of different modalities, leading to suboptimal alignment.

\subsection{Multimodal Generative Tasks}
\textbf{Experimental Settings.} Leveraging the rich cross-domain semantic information and universal topological patterns learned during pre-training, PLANET exhibits strong potential for generative applications.
To address Q3, we conduct evaluations on two distinct tasks:
\ding{182} \textbf{Graph-to-Text (G2Text) Generation.} This task aims to generate a comprehensive textual description for a target node, conditioned on the graph structure and multimodal context. We strictly follow the evaluation settings of MMGL~\cite{yoon2023multimodal}. Specifically, MMGL relies on frozen CLIP encoders, and we replace these embeddings with the text and image embeddings generated by PLANET. 
\ding{183} \textbf{Graph-to-Image (G2Image) Generation.} This task focuses on synthesizing an image for a target node based on a text prompt and neighbor image context. Adopting the framework of InstructG2I~\cite{jin2024instructg2i}, we replace the image features extracted by the pre-trained CLIP image encoder with the topology-enriched image embeddings produced by PLANET.
Implementation details are provided in Appendix~\ref{appendix: Multimodal Generative Tasks}.

\textbf{Results.} Fig.~\ref{fig:g2text} and~\ref{fig:g2image} shows the results of G2Text and G2Image tasks, respectively. 
PLANET consistently outperforms baselines across all metrics.
We attribute this to our Divide-and-Conquer strategy, which produces superior embeddings through modality interaction and alignment.

\subsection{Computational Efficiency Analysis}
\begin{figure}
    \centering
    \includegraphics[width=1.0\linewidth]{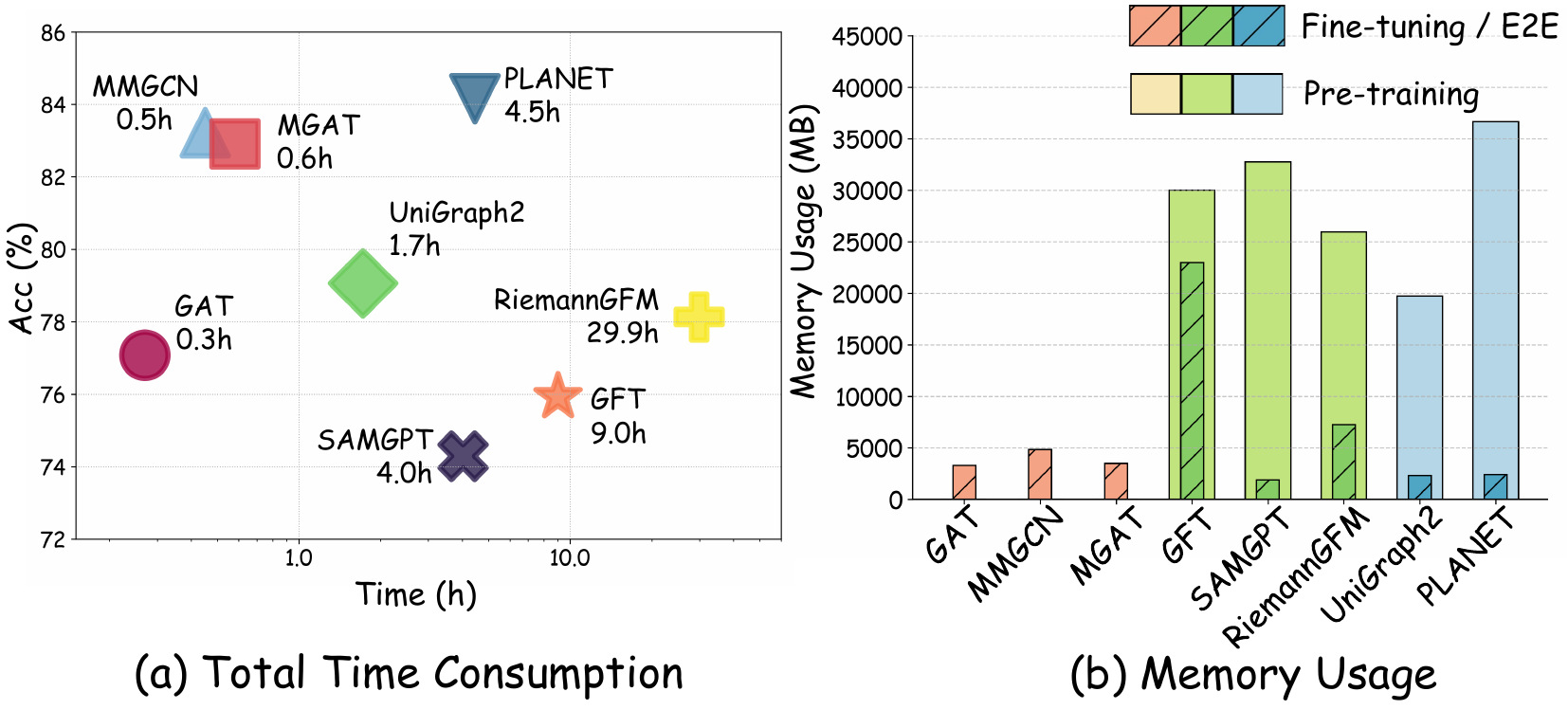}
    \caption{Efficiency comparison between models on Goodreads-NC. (a) Total time consumption. (b) GPU memory usage during different stages.}
    \vspace{-4mm}
    \label{fig:efficiency}
\end{figure}

To answer Q4, we conducted an efficiency analysis on the large-scale Goodreads-NC dataset.
We evaluated the total pre-training and fine-tuning time of various baselines. For task-specific models (i.e., GAT, MMGCN, MGAT), we record their End-to-End (E2E) training time.
Additionally, we monitor the GPU memory usage across different stages.
Results (Fig.~\ref{fig:efficiency}) demonstrate that PLANET achieves a superior efficiency-performance trade-off, validating the scalability of our Divide-and-Conquer design. 
PLANET utilizes sufficient memory for pre-training, its fine-tuning memory usage is minimal since only the linear classifier is trained. 
This indicates that our model enables resource-efficient adaptation for downstream tasks.

\section{Conclusion}


In this work, we propose PLANET to resolve the critical limitations of existing MGFMs in modality interaction and alignment. 
Leveraging a Divide-and-Conquer strategy, the EDG module couples expert-driven extraction with topology-aware attention to facilitate modality interaction at embedding level,
and the NDR module bridges the semantic gap by mapping multimodal representations into a unified Discretized Semantic Representation Space at node level.
Theoretical analysis and extensive experiments confirm the superiority of PLANET, which achieves state-of-the-art performance across diverse graph-centric and multimodal generative tasks, establishing a robust and scalable framework for learning on MAGs.
In future work, we aim to extend PLANET to incorporate richer modalities like audio and video, broadening the applicability of MGFMs in web-scale scenarios.





\newpage
\section*{Impact Statement}
This paper presents work whose goal is to advance the field of Machine
Learning. There are many potential societal consequences of our work, none of 
which we feel must be specifically highlighted here.





\bibliography{icml2026}
\bibliographystyle{icml2026}

\newpage
\appendix
\onecolumn
\section{Datasets}
\textbf{Dataset.} Detailed information of each dataset is presented in Table~\ref{tab:dataset_stats}.
\begin{itemize}
    \item \textbf{MAGB Datasets.} We select four representative graphs from the MAGB benchmark~\cite{yan2025graph}. \textbf{Movies}, \textbf{Toys} and \textbf{Grocery} are e-commerce networks constructed from Amazon. Nodes represent products, and edges indicate also-bought or also-viewed co-purchasing relationships. Each node is associated with a product description and a product image.
    \textbf{Reddit-S} is a social network graph derived from the Reddit platform. 
    Nodes represent user posts containing both text and images.
    Edges connect posts commented on by the same user, reflecting shared user interests.
    \item \textbf{MM-GRAPH Datasets.} We incorporate seven datasets from the MM-GRAPH benchmark~\cite{zhu2025mosaic}.
    \textbf{Amazon-Sports}, \textbf{Amazon-Cloth} and \textbf{Ele-fashion} are e-commerce networks where nodes represent items associated with product titles and images.
    \textbf{Goodreads-NC} and \textbf{Goodreads-LP} are books networks enriched with cover and descriptions. Edges represent user co-interactions, linking books read or liked by the same users.
    \textbf{MM-CoDEx-s} and \textbf{MM-CoDEx-m} are knowledge graphs where entities contain Wikipedia texts and images. Edges denote semantic relations (e.g., born\_in) between entities.
    \item \textbf{Flickr30k.} Flickr30k is a social network. Notably, Flickr30k lacks an intrinsic graph structure. To adapt it for graph learning, we construct a k-NN graph, where edges are established between nodes with high feature similarity to capture latent semantic correlations. Nodes represent images associated with five descriptions. 
\end{itemize}

\textbf{Dataset Split.} For the MAGB datasets and Flickr30k, each dataset is randomly partitioned into training sets (60\%), validation sets (20\%) and testing sets (20\%).
For the MM-GRAPH datasets, we strictly adhere to the official data splits provided in the original benchmark~\cite{zhu2025mosaic}.

\begin{table*}[t]
\centering
\caption{Statistics of MAG datasets.}
\renewcommand\tabcolsep{3.0pt} 
\label{tab:dataset_stats}
\begin{tabular}{ll rr r lrr}
\toprule
\textbf{Domain} & \textbf{Dataset} & \textbf{Avg. \#Nodes} & \textbf{Avg. \#Edges} & \textbf{\#Graphs} & \textbf{Task} & \textbf{\#Classes} & \textbf{\#Pretrain Weights} \\
\midrule
\multirow{6}{*}{ \makecell[l]{E-commerce \\ Network}} & Movies & 16,672 & 218,390 & 1 & NC & 20 & 5.0 \\
 & Toys & 20,695 & 126,886 & 1 & NC & 18 & 5.0 \\
 & Grocery & 17,074 & 171,340 & 1 & NC & 20 & 5.0 \\
 & Amazon-Sports & 50,250 & 356,202 & 1 & LP & -- & 1.0 \\
 & Amazon-Cloth & 125,839 & 951,271 & 1 & LP & -- & 1.0 \\
 & Ele-fashion & 97,766 & 199,602 & 1 & NC & 12 & 1.0 \\
\midrule
\multirow{2}{*}{Social Network} & Reddit-S & 15,894 & 566,160 & 1 & NC & 20 & 10.0 \\
& Flickr30k & 31,783 & 181,551 & 1 & -- & -- & 0.0\\
\midrule
\multirow{2}{*}{Knowledge Graph} & MM-CoDEx-s & 1,383 & 15,884 & 1 & KGC & -- & 20.0 \\
 & MM-CoDEx-m & 7,697 & 52,840 & 1 & KGC & -- & 20.0 \\
\midrule
\multirow{2}{*}{Books Network} & Goodreads-NC & 685,294 & 7,235,048 & 1 & NC & 11 & 0.5 \\
 & Goodreads-LP & 636,502 & 3,437,017 & 1 & LP & -- & 0.5 \\
\bottomrule
\end{tabular}
\end{table*}


\section{Detailed Implementation of Empirical Study}

\subsection{Detailed Implementation of Empirical Study 1}
\label{Appendix_emp_1}
The first empirical study(Fig.~\ref{fig:fig1}(a)\&(b)) is designed to validate the benefits of incorporating multimodal information and the superiority of MGFMs approaches. we conducted a comparative study involving both traditional MGL models (MMGCN, MGAT), Graph Foundation Models (RiemannGFM, SAMGPT) and Multimodal Graph Foundation Models (UniGraph2).

\textbf{Input Modality Settings.}
We standardized the feature dimensions and encoding processes across all settings. 
Specifically, both text data and image data are encoded through Qwen2-VL-7B-Instruct~\cite{wang2024qwen2}. The output features are in a fixed dimension of 3,584. 
For \underline{\textit{Text}} or \underline{\textit{Image}} settings, These settings represent single-modality scenarios. 
we adopted a duplication strategy to ensure compatibility with model architectures which need dual modality inputs. 
For instance, in the \underline{\textit{Text}} setting, we utilize the text embeddings as the primary input and duplicate them to serve as the input for the second modality (i.e., replacing the image features with text features).
For the \underline{\textit{Text+Image}} setting, we provide models with comprehensive multimodal information (i.e., text embeddings and image embeddings).

\textbf{Model Adaptation.}
For models inherently designed to handle MAGs, including MMGCN, MGAT, and UniGraph2, we retain their original architectures without modification.
For GFMs that are not naturally designed to process MAGs, including RiemannGFM and SAMGPT, we concatenated the embeddings from the distinct modalities (e.g., for the \underline{\textit{Text+Image}} setting, we concatenated the text and image embeddings) to form a unified node feature embedding.

\textbf{Results Analysis.}
As shown in Fig.~\ref{fig:fig1}(a)\&(b), the \underline{\textit{Text+Image}} setting consistently outperforms single-modality baselines, corresponding to the conclusion: \ding{182} \textbf{The use of multiple modalities consistently improve model performance.}
Comparing UniGraph2 with RiemannGFM and SAMGPT, UniGraph2 significantly outperforms general GFMs on both dataset across all settings, corresponding to the conclusion: \ding{183} \textbf{MGFMs significantly outperform GFMs.}

\subsection{Detailed Implementation of Empirical Study 2}
\label{Appendix_emp_2}
This empirical study (Fig.~\ref{fig:fig1}(c)\&(d)) is designed to validate the limitations of recent MGFMs, exemplified by the state-of-the-art UniGraph2.

\textbf{Modality Interaction (MI).}
The original UniGraph2 employs an ``early fusion" strategy, where embeddings from different modalities are aggregated via weighted sum before entering graph encoder.
To realize MI, we removed the early fusion mechanism and introduced a naive layer-wise interaction strategy.
Specifically, in each Graph Encoder layer , we allow information to flow from one modality $m'$ to another modality $m$ via a linear projection:
\begin{equation}
\mathbf{h}^{(\ell,m)} = \operatorname{GNNLayer}_{m}^{(l)}\left(\mathbf{h}^{(\ell-1,m)}\right) + \frac{\alpha}{|\Omega|-1} \sum_{m' \neq m}\mathbf{W}_{m'm}^{(l)} \cdot \operatorname{GNNLayer}_{m'}^{(l)}\left(\mathbf{h}^{(\ell-1,m')}\right)
\end{equation}
where $\mathbf{W}_{m'm}^{(l)}$ is a learnable projection matrix in layer $l$, capturing the relationship between modalities $m'$ and $m$. $\alpha$ is a hyperparameter.
 
\textbf{MI+Modality Alignment (MA).}
Building upon the MI, we incorporated the Symmetric InfoNCE Loss~\cite{radford2021learning} to enforce explicit modality alignment. This contrastive objective pulls representations of the same node across different modalities closer in the latent space.


\section{Detailed Formulations}
\label{appendix:detailed_formulations}

\subsection{Modality Masking}
\label{appendix: modality masking}
Inspired by~\citet{hou2022graphmae}, we employ a modality masking strategy.
Formally, for each node $v_i \in \tilde{\mathcal{V}}$ and modality $m \in \Omega$, we have:
\begin{equation}
    \tilde{\mathbf{x}}_i^{(m)} = \mathbf{x}_i^{(m)} \odot \mathbf{b}_i^{(m)},
\end{equation}
where $\mathbf{b}_i^{(m)} \in \{0, 1\}^{d_m}$ is a binary mask vector sampled from a Bernoulli distribution, $\tilde{\mathcal{V}} \subset \mathcal{V}$ is a sampled subset of nodes, and $\odot$ denotes the element-wise product.
This process encourages the model to recover missing feature dimensions using contextual information from the graph and other modalities.

\subsection{Topology-Aware Attention Mechanism}
\label{appendix: EDG}
In Sec.~\ref{sec:interaction}, we abstract the topology-aware attention mechanism as a generic operator $\operatorname{GT}_l(\cdot)$. Here, we provide its detailed mathematical formulation. We strictly follow the implementation of Graph Transformer~\cite{dwivedi2020generalization}, but change the \textit{query},\textit{key} and \textit{value} vectors.
The output of the multi-head attention is computed as:
{
\setlength{\abovedisplayskip}{3pt} 
\setlength{\belowdisplayskip}{3pt} 
\begin{equation}
    \hat{\mathbf{h}}_i^{(\ell,m)} = \mathbf{O}^{(\ell,m)} \mathop{\Big\|}_{k=1}^{H} \left(\sum_{j \in \mathcal{N}_i} \alpha_{ij,k}^{(\ell,m)} \mathbf{V}^{(\ell,m)}_k \mathbf{e}_{j}^{(\ell,m)} \right),
\end{equation}}%
where $H$ represents the number of heads. $\|$ denotes the concatenation operation, $\mathcal{N}_i$ denotes the set of incoming neighbors for node $i$, and $\mathbf{O}^{(\ell,m)}$ is the output projection matrix. For the $k$-th head, $\mathbf{V}^{(\ell,m)}_k$ maps the expert-distilled neighbor signal $\mathbf{e}_{j}^{(\ell,m)}$ to the \textit{value} vector. The attention coefficient $\alpha_{ij,k}^{(\ell,m)}$ determines the importance of neighbor $j$'s cross-modal information to node $i$:
{
\setlength{\abovedisplayskip}{3pt} 
\setlength{\belowdisplayskip}{3pt} 
\begin{equation}
    \alpha_{ij,k}^{(\ell,m)} = \text{softmax}_j \left( \frac{\left(\mathbf{Q}^{(\ell,m)}_k \mathbf{h}_i^{(\ell-1,m)}\right)^{\top} \left(\mathbf{K}^{(\ell,m)}_k \mathbf{e}_j^{(\ell,m)}\right)}{\sqrt{d_k}} \right),
\end{equation}}%
where $d_k$ is the dimension of each head. The projection matrices $\mathbf{Q}^{(\ell,m)}_k, \mathbf{K}^{(\ell,m)}_k$ transform the input features into \textit{query} and \textit{key} vectors, respectively. 
After that, a residual connection followed by Layer Normalization~\cite{ba2016layer} is applied to the output of the attention mechanism:
\begin{equation}
    \hat{\hat{\mathbf{h}}}_i^{(\ell,m)} = \operatorname{LayerNorm}\left(\mathbf{h}_i^{(\ell-1,m)} + \hat{\mathbf{h}}_i^{(\ell,m)}\right).
\end{equation}
The normalized representation is then passed through a Feed-Forward Network (FNN):
\begin{equation}
    \hat{\hat{\hat{\mathbf{h}}}}_i^{(\ell,m)} = \operatorname{MLP}_{\ell}^{2} \left(\operatorname{ReLU\left(\operatorname{MLP}^{1}_{\ell}\left(\hat{\hat{\mathbf{h}}}_i^{(\ell,m)}\right)\right)}\right).
\end{equation}
Finally, a second residual connection and Layer Normalization are applied to produce the final node embedding for layer $l$:
\begin{equation}
    \mathbf{h}_i^{(\ell,m)} = \operatorname{LayerNorm}\left(\hat{\hat{\mathbf{h}}}_i^{(\ell,m)}+\hat{\hat{\hat{\mathbf{h}}}}_i^{(\ell,m)}\right).
\end{equation}

\subsection{Loss Functions}
\label{appendix: loss}

\textbf{MoE Load Balancing.} 
To prevent the issue of expert collapse (i.e., a small subset of experts dominates the processing while others remain idle), we consider the MoE load balancing loss. 
Adopting the standard formulation proposed by \citet{fedus2022switch}:
\begin{equation}
    \mathcal{L}_{load} = K \cdot \sum_{k=1}^{K} P_k f_k,
\end{equation}
where $f_k$ is the fraction of samples assigned to expert $k$, and $P_k$ denotes the average routing probability for expert $k$ across a batch.

\textbf{Structural Reconstruction.}
To preserve topological information, we minimize the binary cross-entropy loss over observed edges $\mathcal{E}$ (positive samples) and randomly sampled unconnected pairs $\hat{\mathcal{E}}$ (negative samples):
\begin{equation}
\begin{aligned}
    \mathcal{L}_{topo} =  \sum_{m \in \Omega} \Bigg[ -\frac{1}{|\Omega||\mathcal{E}|} \sum_{(i,j)\in \mathcal{E}} \log\left(\sigma\left({\mathbf{u}_i^{(m)}}^\top \cdot \mathbf{u}_j^{(m)}\right)\right)
    -\frac{1}{|\Omega||\hat{\mathcal{E}}|} \sum_{(i',j') \in \hat{\mathcal{E}}} \log\left(1-\sigma\left({\mathbf{u}_{i'}^{(m)}}^\top \cdot \mathbf{u}_{j'}^{(m)}\right)\right) \Bigg],\\
\end{aligned}
\end{equation}
where $\operatorname{SRD}$ is a structural projection head, $\sigma$ is the sigmoid function, and $\mathbf{u}_i^{(m)} = \mathcal{D}_m^{SR}(\mathbf{h}_i^{(all,m)})$, $\mathcal{D}_{m}^{SR}$ is structural reconstruction decoder for modality $m$, which is an MLP.

\section{Proof of Theoretical Analysis}
\label{appendix: proof}

\subsection{Proof of Theorem~\ref{thm:synergy}}

In this section, we utilize the Information Bottleneck (IB) principle to prove that vanilla Multimodal Graph Encoders (e.g., MMGCN) inevitably discard synergistic features, while PLANET's topology-aware interaction preserves them.

Following the formulation in Graph Information Bottleneck~\cite{wu2020graph}, the goal of graph representation learning is to maximize the IB Lagrangian $\mathcal{L}_{IB}(Z) = I(Y; Z) - \beta I(X; Z)$, where $\beta > 0$ controls the compression trade-off. 

\textbf{Vanilla Multimodal Graph Encoder.} The Vanilla Multimodal Graph Encoder processes modalities independently. For a specific modality $A$, the encoding follows the Markov chain $Y \leftrightarrow G^{(A)} \rightarrow Z_A$. The optimization objective is to maximize the local IB: $\mathcal{L}_{Vanilla} = I(Y; Z_A) - \beta I(G^{(A)}; Z_A)$. To prove that this encoder discards synergistic features, We define two potential encoding strategies:
\begin{itemize}
    \item \textbf{Drop Synergistic Features (DSF):} The encoder captures only unique features, i.e., $Z_A^- \approx \{U_A\}$.
    \item \textbf{Keep Synergistic Features (KSF):} The encoder captures both unique and synergistic features, i.e., $Z_A^+ \approx \{U_A, S_A\}$.
\end{itemize}
We calculate the Marginal Contribution $\Delta \mathcal{L}_{Vanilla}$ of shifting from Strategy DSF to Strategy KSF, which can be expressed as $\Delta \mathcal{L}_{Vanilla} = \Delta I_Y - \beta \cdot \Delta I_{G^{(A)}}$.
For $\Delta I_Y$, we apply the Chain rule of mutual information~\cite{federici2020learning}:
\begin{equation}
\begin{aligned}
    \Delta I_Y &= I(Y; Z_A^+) - I(Y; Z_A^-) \\
    &= I(Y; U_A, S_A) - I(Y; U_A) \\
    &= I(Y; U_A) + I(Y; S_A \mid U_A) - I(Y; U_A) \\
    &= I(Y; S_A \mid U_A).
\end{aligned}
\end{equation}
Crucially, under the independent encoding setting, modality $B$ is unobservable. According to Definition~\ref{def:synergy}, the synergistic feature $S_A$ contains no information about $Y$ without the presence of $S_B$ (i.e., $I(Y; S_A \mid U_A) \approx 0$). Therefore, we have $\Delta I_Y \approx 0$.

For $\Delta I_{G^{(A)}}$, we similarly apply the Chain rule:
\begin{equation}
\begin{aligned}
    \Delta I_{G^{(A)}} &= I(G^{(A)}; Z_A^+) - I(G^{(A)}; Z_A^-) \\
    &= I(G^{(A)}; U_A, S_A) - I(G^{(A)}; U_A) \\
    &= I(G^{(A)}; S_A \mid U_A)\\
    &= H(S_A \mid U_A) - H(S_A \mid G^{(A)}, U_A).
\end{aligned}
\end{equation}
Since $S_A$ is intrinsically part of the input $G^{(A)}$, we have $H(S_A \mid G^{(A)}, U_A)=0$, therefore:
\begin{equation}
    \Delta I_{G^{(A)}} = H(S_A \mid U_A) > 0.
\end{equation}

Substituting these results back to the Marginal Contribution $\Delta \mathcal{L}_{Vanilla}$:
\begin{equation}
    \Delta \mathcal{L}_{Vanilla} \approx 0 - \beta \cdot H(S_A \mid U_A) < 0.
\end{equation}
Since the contribution is negative for any $\beta > 0$, maximizing the local IB objective necessitates the exclusion of synergistic features to avoid incurring unnecessary compression costs. Consequently, we have $Z^*_{Vanilla} \approx \{U_A, U_B\}$.

\textbf{PLANET with the EDG Module.} The topology-aware interaction mechanism aggregates semantic context from the cross-modal neighborhood. This implies that the encoding process follows a unified Markov chain $Y \leftrightarrow \{G^{(A)}, G^{(B)}\} \rightarrow Z_{EDG}$. The optimization objective is to maximize the local IB: $\mathcal{L}_{EDG} = I(Y; Z_{EDG}) - \beta I(G^{(A)},G^{(B)}; Z_{EDG})$. 

Similarly, we contrast two strategies: The first strategy captures only unique features, i.e., $Z_{EDG}^-=\{U_A, U_B\}$, and the second strategy additionally captures the synergistic pair, i.e., $Z_{EDG}^+=\{U_A,U_B,S_A, S_B\}$. The marginal contribution of transitioning from the first strategy to the second strategy is formulated as:
\begin{equation}
    \Delta \mathcal{L}_{EDG} = \Delta I_Y - \beta \cdot \Delta I_{\{G^{(A)}, G^{(B)}\}}.
\end{equation}

Applying the chain rule yields $\Delta I_{\{G^{(A)},G^{(B)}\}} = H(S_A, S_B \mid U_A, U_B) > 0$ and $\Delta I_Y = I(Y; S_A, S_B \mid U_A, U_B)$. 
Under the joint view enabled by the unified Markov chain, the synergistic features become informative. By Definition~\ref{def:synergy}, the interaction information is strictly positive, yielding:
\begin{equation}
    \Delta I_Y = I(Y; S_A, S_B \mid U_A, U_B) \gg 0.
\end{equation}
Consequently, we have $\Delta \mathcal{L}_{EDG} = \Delta I_Y - \beta \cdot H(S_A, S_B \mid U_A, U_B)$. 
Given that the significant synergistic information gain outweighs the weighted compression cost (i.e., $\Delta I_Y > \beta \cdot \Delta I_{\{G^{(A)}, G^{(B)}\}}$), $\Delta \mathcal{L}_{EDG}$ becomes strictly positive. Consequently, maximizing the joint IB objective necessitates the preservation of these features, leading to $Z^*_{EDG} \approx \{U_A, U_B, S_A, S_B\}$.

Finally, we quantify the information gap between the two representations:
\begin{equation}
    I(Y; Z^*_{EDG}) - I(Y; Z^*_{Vanilla}) \approx I(Y; U_A, U_B, S_A, S_B) - I(Y; U_A, U_B) = I(Y; S_A, S_B \mid U_A, U_B) > 0.
\end{equation}
This theorem demostrates that our EDG module captures a strictly larger amount of relevant information than vanilla Multimodal Graph Encoders which process MAGs as independent graphs.

\subsection{Proof of Theorem~\ref{thm:alignment}}
To analyze the generalization bound of our alignment strategy, we employ the 1-Wasserstein distance $W_1(\hat{\mu}_m, \hat{\mu}_t)$ as the metric for distributional discrepancy.

Utilizing the triangle inequality property of the Wasserstein metric, we decompose the total alignment error into three components:
\begin{equation}
\label{eq: triangle}
    W_1(\hat{\mu}_m, \hat{\mu}_t) \le W_1(\hat{\mu}_m, \hat{\nu}_m) + W_1(\hat{\nu}_m, \hat{\nu}_t) + W_1(\hat{\nu}_t, \hat{\mu}_t).
\end{equation}

\textbf{For $W_1(\hat{\mu}_m, \hat{\nu}_m)$.} 
By the Kantorovich formulation of Optimal Transport~\cite{villani2008optimal}, the Wasserstein distance is defined as the infimum of transport costs over all valid joint couplings $\Pi(\hat{\mu}_m, \hat{\nu}_m)$:
\begin{equation}
    W_1(\hat{\mu}_m, \hat{\nu}_m) = \inf_{\pi \in \Pi(\hat{\mu}_m, \hat{\nu}_m)} \mathbb{E}_{(x, y) \sim \pi} [\|x - y\|_2].
\end{equation}
We define $\pi^*$ such that it transports the probability mass $1/N$
associated with each sample $x_i^{(m)}$ directly to its corresponding quantized token $Q(x_i^{(m)})$. 
Formally, the joint distribution $\pi^*$ is supported exclusively on the set of pairs $\{(x_i^{(m)}, Q(x_i^{(m)}))\}_{i=1}^N$. 
Let $\mathcal{C}(\pi)$ denote the transport cost associated with a coupling $\pi$. Under our deterministic coupling $\pi^*$, this cost is exactly:
\begin{equation}
    \mathcal{C}(\pi^*) = \frac{1}{N} \sum_{i=1}^N \|x_i^{(m)} - Q(x_i^{(m)})\|_2 = \mathbb{E}_{x \sim \hat{\mu}_m} \|x - Q(x)\|_2.
\end{equation}
Therefore, we have the inequality:
\begin{equation}
\label{eq: mum,vm}
    W_1(\hat{\mu}_m, \hat{\nu}_m) \le \mathcal{C}(\pi^*) = \mathbb{E}_{x \sim \hat{\mu}_m} \|x - Q(x)\|_2.
\end{equation}

\textbf{For $W_1(\hat{\nu}_t, \hat{\mu}_t)$.} 
Similarly, due to the symmetry of the Wasserstein metric and the same quantization mechanism applied to the anchor text modality $t$, we have the corresponding upper bound:
\begin{equation}
\label{eq: mut,vt}
    W_1(\hat{\nu}_t, \hat{\mu}_t) = W_1(\hat{\mu}_t, \hat{\nu}_t) \le \mathbb{E}_{x \sim \hat{\mu}_t} \|x - Q(x)\|_2.
\end{equation}

\textbf{For $W_1(\hat{\nu}_m, \hat{\nu}_t)$.} 
We denote $\nu^*_m$ and $\nu^*_t$ as the underlying ground-truth discrete distributions for modality $m$ and the anchor text modality $t$, both supported on $\mathcal{S}$. Accordingly, the empirical push-forward measures $\hat{\nu}_m$ and $\hat{\nu}_t$ are viewed as samples drawn from these corresponding ground-truth distributions.
Leveraging the triangle inequality, we have:
\begin{equation}
    W_1(\hat{\nu}_m, \hat{\nu}_t) \le W_1(\hat{\nu}_m, \nu_m^*) + W_1(\nu^{*}_m,\nu^*_t) + W_1(\nu_t^*, \hat{\nu}_t).
\end{equation}
Let $\delta \in (0, 1)$ be the scale parameter. According to Proposition 1 in \citet{weed2019sharp}, The 1-Wasserstein distance is bounded by:
\begin{equation}
\label{eq: bound}
     \mathbb{E} [W_1(\hat{\nu}_m, \nu^*_m)] \lesssim \delta^{k^*} + \sum_{k=1}^{k^*} \delta^{k-1} \sum_{Q_i^{k} \in Q^{k}} \mathbb{E} \left[| \hat{\nu}_m(Q_i^{k}) - \nu^*_m(Q_i^{k}) | \right],
\end{equation}
where $k^*$ is the truncation depth, and $\mathcal{Q}^{k}$ is the partition at scale $k$. 

We simplify the bound by observing three key properties: 
\ding{172} For a sufficiently large truncation depth $k^*$, the partition resolution exceeds the minimum separation of DSRS vectors, thereby eliminating the truncation error (i.e., $\delta^{k^*} \to 0$). 
\ding{173} The finite support restricts the inner summation $\sum_{Q_i^{k} \in Q^{k}}$ to at most $C$ non-zero terms. 
\ding{174} The term $\mathbb{E} [ | \hat{\nu}_m(Q_i^{k}) - \nu^*_m(Q_i^k) | ]$ corresponds to the mean absolute deviation of empirical frequencies. By Jensen's inequality and the variance bound of the binomial distribution, this is strictly bounded by $\sqrt{\frac{1}{4N}} = \frac{1}{2\sqrt{N}}$.

Substituting these properties into Eq.~\eqref{eq: bound}:
\begin{equation}
\begin{aligned}
    \mathbb{E} [W_1(\hat{\nu}_m, \nu^*_m)] &\le \sum_{k=1}^{\infty} \delta^{k-1} \sum_{Q_i^{k} \in \mathcal{Q}^k} \mathbb{E} \left[| \hat{\nu}_m(Q_i^{k}) - \nu^*_m(Q_i^{k}) | \right] \\
    &\le \sum_{k=1}^{\infty} \delta^{k-1} \cdot \left( \sum_{j=1}^C \frac{1}{2\sqrt{N}} \right) \\
    &= \left( \sum_{k=1}^{\infty} \delta^{k-1} \right) \cdot \frac{C}{2\sqrt{N}}.
\end{aligned}
\end{equation}

Since $\delta \in (0, 1)$, the geometric series converges to a constant $C_\delta$, yielding $\mathbb{E} [W_1(\hat{\nu}_m, \nu^*_m)] \le C_\delta \cdot \frac{C}{\sqrt{N}} = O\left( \frac{C}{\sqrt{N}} \right)$.
Both $W_1(\hat{\nu}_m, \nu^*_m)$ and $W_1(\nu^*_t, \hat{\nu}_t)$ satisfy this bound, the total discrete alignment error converges at the rate:
\begin{equation}
\label{eq: vm,vt}
    \mathbb{E}[W_1(\hat{\nu}_m, \hat{\nu}_t)] \le \mathbb{E}[W_1(\hat{\nu}_m, \nu^*_m)] + \mathbb{E}[W_1(\nu^*_m, \nu^*_t)] + \mathbb{E}[W_1(\nu^*_t, \hat{\nu}_t)] \le \mathbb{E}[W_1(\nu^*_m, \nu^*_t)] + O\left(\frac{C}{\sqrt{N}}\right)
\end{equation}

Substituting Eq.~\eqref{eq: mut,vt},~\eqref{eq: vm,vt},~\eqref{eq: mum,vm} into Eq.~\eqref{eq: triangle}, we obtain the final bound:
\begin{equation}
    W_1(\hat{\mu}_m, \hat{\mu}_t) \le \mathbb{E}_{x \sim \hat{\mu}_m} \|x - Q(x)\|_2 + \mathbb{E}_{z \sim \hat{\mu}_t} \|z - Q(z)\|_2 + W_1(\nu^*_m,\nu^*_t) + O\left(\frac{C}{\sqrt{N}}\right).
\end{equation}

\textbf{Comparison.} 
Consider the standard case in continuous space $\mathbb{R}^d$. According to Theorem 1 and Proposition 7 in \citet{weed2019sharp}, the convergence rate of Wasserstein estimation is governed by the lower Wasserstein dimension $d_*(\mu)$ of the measure. For continuous feature distributions in high-dimensional spaces (e.g., dense semantic embeddings in $\mathbb{R}^d$), the intrinsic dimension $d_*(\mu)$ typically equals the ambient dimension $d$. Consequently, we have:
\begin{equation}
\label{eq: comparison}
    \mathbb{E}[W_1(\hat{\mu}, \mu)] \gtrsim N^{-1/d}.
\end{equation}

\textbf{Conclusion.}
Eq.~\eqref{eq: comparison} implies that for high-dimensional features, the convergence rate is extremely slow.
In contrast, by leveraging the DSRS where modality alignment is explicitly constrained by the General Knowledge Loss, PLANET accelerates the convergence rate to $O(C/\sqrt{N})$. Simultaneously,
Minimizing $\mathcal{L}_{VQ}$ in Eq.~\eqref{eq: VQ loss} effectively minimizes the upper bounds of $W_1(\hat{\mu}_m, \hat{\nu}_m)$ and $W_1(\hat{\nu}_t, \hat{\mu}_t)$, ensuring that continuous features lie close to DSRS.
These mechanisms achieve effective modality alignment.

\section{Experiment Settings}
\label{experiment settings}

\subsection{Supervised Learning}
\label{appendix: supervised learning}

\textbf{Baseline Settings.}
For supervised models (i.e., GCN, MMGCN, MGAT), we train them directly on downstream datasets without any pre-training.
In contrast, for self-supervised and graph foundation models, we follow the pre-training and fine-tuning paradigm: models are first pre-trained across various datasets and subsequently fine-tuned on each specific downstream task.

During the fine-tuning stage, for baselines that do not provide specific methods for node classification or link prediction, we add an MLP to perform these tasks.
Specifically, the input embeddings of MLP are encoded by the pretrained graph encoder. 
For node classification, we set the output dimension of the MLP to the number of classes. 
For link prediction, we concatenate the embeddings of the target node pair as input to the MLP, where the output dimension is 1, reflecting the score for predicting the pair as a positive sample.

\textbf{Feature Encoding.}
Both raw text and image data are encoded using Qwen2-VL-7B-Instruct~\cite{wang2024qwen2} to generate high-quality initial node embeddings.
For baselines that are incapable of processing MAGs (e.g., GraphMAE2, GFT), the embeddings from modality-specific encoders are concatenated along the feature dimension.
It is worth noting that while RiemannGFM and SAMGPT prioritize the learning of topological knowledge, they use node features in specific ways (e.g., RiemannGFM uses node features in downstream tasks). In our implementation, we replace these node features with the representations encoded by the modality-specific encoders, which have significantly richer semantic information.

\textbf{Pre-training Datasets and Sampling Strategy.}
Self-supervised and foundation models are pre-trained in same weights of datasets (weights are presented in Table~\ref{tab:dataset_stats}).
For baselines that are incapable of taking multiple datasets as input or cannot scale to large-scale graphs, we implement a sampling approach. 
Specifically, for each node in the graph, we extract its 2-hop neighbor subgraph. We then randomly sample a fixed number of these subgraphs from each dataset according to the pre-defined weights (Table~\ref{tab:dataset_stats}), forming the training batches. This guarantees that all models are pre-trained on an identical data distribution.

\subsection{Few-Shot Learning}
\textbf{Overall Settings.}
Most of our experimental settings are the same as those in Appendix~\ref{appendix: supervised learning} (e.g., feature encoding, sampling strategies.)
Following~\citet{wang2024gft}, we adapt the "Pre-training and Fine-tuning" paradigm. However, during the fine-tuning stage, we do not use an MLP as the classification head. 
Instead, we use a prototype-based method for classification.
The specific values for the parameters used in our experiments are detailed in Table~\ref{tab:few-shot parameter settings}.

\begin{table*}[t]
\centering
\caption{Detailed parameter settings for few-shot node classification and link classification tasks.}
\label{tab:few-shot parameter settings}
\begin{tabular}{llc}
\toprule
\textbf{Task} & \textbf{Parameter} & \textbf{Value} \\
\midrule
\multirow{3}{*}{Few-shot Node Classification} & $n\_train$ & 20 \\
 & $n\_query$ & 10 \\
 & $n\_task$ & 10 \\
\midrule 
\multirow{3}{*}{Few-shot Link Classification} & $n\_train$ & 20 \\
 & $n\_query$ & 40 \\
 & $n\_task$ & 10 \\
\bottomrule
\end{tabular}
\end{table*}

\textbf{Few-Shot Node Classification.}
For an $N$-way $K$-shot task, we randomly sample $n\_train$ samples for each class from the training set.
On the test set, we randomly select $N$ classes and sample $K$ instances per class to serve as prototype vectors, along with $n\_query$ instances for evaluation. 
These $N \cdot K$ samples are used to construct a prototype classifier, where the embedding of each class is computed by averaging its corresponding $K$ sample embeddings, resulting in $N$ class embeddings
For evaluation, we calculate the cosine similarity between the evaluation vector and each class embedding, assigning the sample to the class with the highest similarity.
The same procedure applies to the validation set. To eliminate randomness, this sampling process is repeated $n\_task$ times, with the final result being the average of these runs.

\textbf{Few-Shot Link Classification.}
Since the original datasets (i.e., Amazon-Sports, Amazon-Cloth) are designed for link prediction tasks, we adapt them for link classification tasks. 
Specifically, we aggregate all positive edges from the training, validation, and test sets to form positive samples.
We simultaneously generate an equivalent number of random negative edges to serve as negative samples. 
These negative edges are allocated to the training, validation, and test sets to strictly match the count of positive edges in each respective set (e.g., a test set containing 500 positive edges is assigned 500 negative edges). 
Consequently, we formulate the problem as a balanced $2$-way $K$-shot binary classification task.
The subsequent method follows the \textbf{Few-shot Node Classification} described above.

\subsection{Multimodal Generative Tasks}
\label{appendix: Multimodal Generative Tasks}
\textbf{G2Text.} We evaluate the performance of G2Text generation on two multimodal datasets:
\begin{itemize}
    \item \textbf{Grocery.} This is a real-world e-commerce dataset. The objective is to generate a comprehensive product description for a target node. The input consists of the target node's product title and the multimodal context (images and texts) from its neighbors (including target node's image). 
    \item \textbf{Flickr30k.} Originally an image-text retrieval dataset, we adapt it for graph-based generation by constructing a k-NN graph based on feature similarity. Each node contains an image and five distinct textual descriptions. For the G2Text task, we utilize the last four descriptions of the target node and the multimodal information of neighbor nodes to generate the first description. 
\end{itemize}
For the specific prompts used for each dataset, please refer to Table~\ref{tab: prompts}.

\begin{table*}[t]
\centering
\renewcommand{\arraystretch}{1.1} 
\setlength{\aboverulesep}{0pt}
\setlength{\belowrulesep}{0pt}
\caption{Specific prompts for Grocery and Flickr30k datasets in G2Text task.}
\label{tab: prompts}
\begin{tabular}{l | l}
\toprule
\multicolumn{1}{c|}{\textbf{Dataset}} & \multicolumn{1}{c}{\textbf{Prompt}} \\
\midrule
\multirow{3}{*}{Grocery} & \#\#\# Task: Generate a natural-language description of the product node. \\
& \#\#\# Input: Title: \textit{\textless item title of the target node \textgreater}. \\
& \#\#\# Output Results: \\
\midrule 
\multirow{3}{*}{Flickr30k} & \#\#\# Task: Generate a detailed description for the image based on the context. \\
& \#\#\# Input: Context: \textit{\textless desc 2 \textgreater}, \textit{\textless desc 3 \textgreater}, \textit{\textless desc 4 \textgreater}, \textit{\textless desc 5 \textgreater}. \\
& \#\#\# Output Results: \\
\bottomrule
\end{tabular}
\end{table*}

\textbf{G2Image.} We evaluate the G2Image task on two datasets: Goodreads-NC and Ele-fashion. Following the experimental setting of InstructG2I~\cite{jin2024instructg2i}, we focus on specific, visually distinct categories to verify whether our embeddings capture sufficient semantic nuances to drive fine-grained generation. 
For Goodreads-NC, we select nodes belonging to the history and children categories for training and testing.
For Ele-fashion, we select nodes from the jewelry and shoes categories.
Distinct from the original data partition used in InstructG2I, we adopt a standard randomized partition strategy for our experiments. Specifically, for each selected category subset, we split the data into a training set (80\%) and a testing set (20\%). This ratio ensures that the model is trained on sufficient data while reserving a substantial portion for robust evaluation.

\section{Implementation Notes}
Detailed hyper-parameter settings for the pre-training stage are summarized in Table~\ref{tab:hyperparams}. We optimize the entire framework using AdamW optimizer for 5 epochs.

\begin{table}[t]
    \centering
    \caption{Detailed hyper-parameter settings for pre-training.}
    \label{tab:hyperparams}
    \begin{tabular}{lcc}
        \toprule
        \textbf{Category} & \textbf{Parameter} & \textbf{Value} \\
        \midrule
        \multirow{5}{*}{Optimization} 
        & learning rate & $4\text{e}-5$ \\
        & weight decay & $5\text{e}-4$ \\
        & batch size & 128 \\
        & epochs & 5 \\
        & optimizer & AdamW \\
        \midrule
        \multirow{4}{*}{EDG Module} 
        & num\_experts & 5 \\
        & top-$k$ & 2 \\
        & num\_layers & 8 \\
        & num\_heads & 8 \\
        \midrule
        \multirow{3}{*}{NDR Module} 
        & DSRS\_size & 20480 \\
        & DSRS\_dim & 768 \\
        & DSRS\_temperature & 0.93 \\
        \midrule
        \multirow{6}{*}{Loss} 
        & $\beta_{1}$ (feature reconstruction) & 0.1 \\
        & $\beta_{2}$ (structure reconstruction) & 0.1 \\
        & $\beta_{3}$ (general knowledge) & 0.2 \\
        & $\beta_{4}$ (VQ) & 0.1 \\
        & $\beta_{5}$ (MoE load balance) & 0.01 \\
        & $\beta_{inter}$ & 0.5 \\
        \midrule
        \multirow{5}{*}{Other Params} 
        & hidden\_dim ($d$) & 768 \\
        & dropout & 0.1 \\
        & select\_node\_mask\_p & 0.6 \\
        & select\_modality\_mask\_p & 0.4 \\
        & edge\_mask\_p & 0.15 \\
        \bottomrule
    \end{tabular}
\end{table}


\end{document}